%% file: acl_latex.tex
\pdfoutput=1
\documentclass[11pt]{article}

\usepackage[final]{acl}
\usepackage{times}
\usepackage{latexsym}
\usepackage{xcolor}
\usepackage{todonotes}

\usepackage[T1]{fontenc}
\usepackage[utf8]{inputenc}

\usepackage{microtype}

\usepackage{inconsolata}

\usepackage{graphicx}
\usepackage{amsmath}
\usepackage{amssymb}
\usepackage{amsfonts}
\usepackage{booktabs}
\usepackage{lipsum}
\usepackage{multicol}
\usepackage{multirow}
\usepackage{bm}
\usepackage{cleveref}
\usepackage{easyeqn}
\usepackage{listings}
\usepackage{placeins}
\usepackage{color,xcolor,colortbl}
\usepackage[multiple]{footmisc}
\usepackage{paralist}
\usepackage{subcaption}
\usepackage{enumitem}

\usepackage{graphicx}

\newcommand{\xv}{\mathbf{x}}
\newcommand{\qv}{\mathbf{q}}
\newcommand{\tv}{\mathbf{t}}
\newcommand{\yv}{\mathbf{y}}

\newcommand{\TC}{\mathcal{T}}
\newcommand{\EC}{\mathcal{E}}
\newcommand{\QC}{\mathcal{Q}}
\newcommand{\DC}{\mathcal{D}}

\newcommand{\XSet}{\mathcal{X}}
\newcommand{\uv}{\mathbf{u}}
\newcommand{\multirowcell}[1]{\begin{tabular}[c]{@{}c@{}}#1\end{tabular}}

\title{Token-Level Density-Based Uncertainty Quantification Methods for Eliciting Truthfulness of Large Language Models}

\author{
Artem Vazhentsev\textsuperscript{1,2} \quad
Lyudmila Rvanova\textsuperscript{2,5} \quad
Ivan Lazichny\textsuperscript{2,4} \enspace
\\
\bf Alexander Panchenko\textsuperscript{1,2}\quad
Maxim Panov\textsuperscript{3}\quad
Timothy Baldwin\textsuperscript{3,6}\quad
Artem Shelmanov\textsuperscript{3}\\
\textsuperscript{1}Skoltech \; 
\textsuperscript{2}AIRI \;
\textsuperscript{3}MBZUAI \;
\textsuperscript{4}MIPT \; \\
\textsuperscript{5}FRC CSC RAS \;
\textsuperscript{6}The University of Melbourne \; 
\\
\href{mailto:vazhentsev@airi.net}{vazhentsev@airi.net} ~~ 
\href{mailto:artem.shelmanov@mbzuai.ac.ae}{artem.shelmanov@mbzuai.ac.ae}
}

\begin{document}
\maketitle
\begin{abstract}
  Uncertainty quantification (UQ) is a prominent approach for eliciting truthful answers from large language models (LLMs). To date, information-based and consistency-based UQ have been the dominant UQ methods for text generation via LLMs. Density-based methods, despite being very effective for UQ in text classification with encoder-based models, have not been very successful with generative LLMs. In this work, we adapt Mahalanobis Distance (MD) -- a well-established UQ technique in classification tasks -- for text generation and introduce a new supervised UQ method. Our method extracts token embeddings from multiple layers of LLMs, computes MD scores for each token, and uses linear regression trained on these features to provide robust uncertainty scores. Through extensive experiments on eleven datasets, we demonstrate that our approach substantially improves over existing UQ methods, providing accurate and computationally efficient uncertainty scores for both sequence-level selective generation and claim-level fact-checking tasks. Our method also exhibits strong generalization to out-of-domain data, making it suitable for a wide range of LLM-based applications.\footnote{The code is available online at \url{https://github.com/ArtemVazh/token_mahalanobis_distance}}
\end{abstract}

\section{Introduction}

  \begin{figure*}[t!]
    \centering
    \includegraphics[trim={0.cm 0.cm 0.cm 0.cm},clip,width=1.\linewidth]{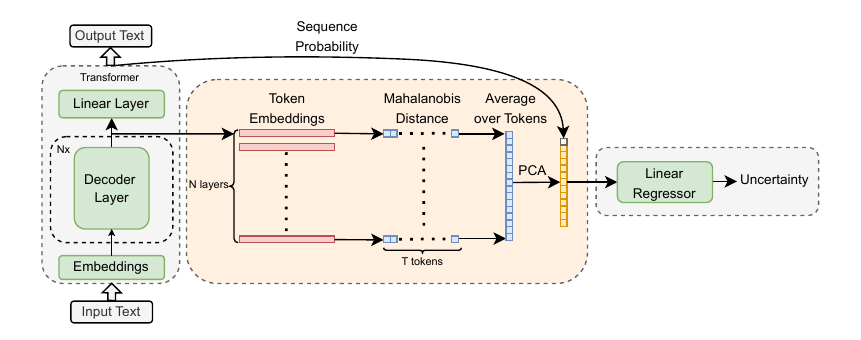}
    \caption{
    An illustration of the proposed method. After each decoder layer, the embeddings of each generated token are extracted. Subsequently, we compute the Mahalanobis distance for each token and layer and then average over all tokens in the generated sequence. Finally, we train a linear regression on the PCA decomposition of the calculated features with the addition of sequence probability to predict the uncertainty of the generation.}
    \label{fig:tmd_scheme}
  \end{figure*}

  Large language models (LLMs) have achieved impressive results over various tasks and applications~\cite{openai2024gpt4technicalreport,llama3,gemma2}. Nevertheless, even the most advanced LLMs are inevitably prone to making mistakes during text generation. Their responses often contain hallucinations or non-factual claims~\cite{xiao-wang-2021-hallucination,dziri-etal-2022-origin}, posing significant challenges for LLM deployment in safety-critical domains. 

  Many studies have investigated methods for assessing the truthfulness of LLM responses~\cite{manakul2023selfcheckgpt,min2023factscore,chen2023hallucination,feng-etal-2024-dont}. However, many of the proposed techniques have limited practical applicability, as they often rely on external knowledge sources or require ensembling multiple large LLMs, leading to high computational costs that make them economically unfeasible for many use cases.

  One of the most promising approaches to addressing this challenge is uncertainty quantification (UQ)~\cite{gal2016dropout,shelmanov-etal-2021-certain,baan2023uncertainty,geng2023survey,fadeeva2023lm}. This research direction recognizes that we will never have complete information about model predictions due to the limited amount of training data and ambiguity of the tasks, and suggests general ways to estimate the reliability of predictions under different conditions.
  Recently, a suite of UQ methods specifically designed for text generation with LLMs has been developed~\cite{fomicheva-etal-2020-unsupervised,lin2023generating,kuhn2023semantic,farquhar2024detecting,duan-etal-2024-shifting}. However, many of these methods are either ineffective or come with a substantial computational overhead, limiting their practicality for large-scale or real-time applications.

  For text classification and regression tasks, researchers have identified several groups of techniques that maintain a balance between effectiveness and computational efficiency~\cite{zhang-etal-2019-mitigating,he2020towards,xin-etal-2021-art,wang-etal-2022-uncertainty,vazhentsev-etal-2023-hybrid,he-etal-2024-uncertainty}. One such class of approach is so-called \textit{density-based uncertainty scores}~\cite{lee2018simple,ddu_amersfoort,nuq_kotelevskii,yoo-etal-2022-detection}. These methods use embeddings of instances obtained from the top layers of a classification model to fit the density of the training distribution in the latent space. The likelihood of the input data under this estimated distribution is then used for confidence estimation.
  This has been demonstrated to achieve excellent results in out-of-distribution detection tasks~\cite{podolskiy2021revisiting}, and proven to be useful for selective text classification~\cite{vazhentsev-etal-2022-uncertainty,vazhentsev-etal-2023-hybrid}. 
  Despite being computationally lightweight, these techniques often outperform more resource-intensive methods, such as deep ensembles~\cite{NIPS2017_9ef2ed4b} and Monte Carlo dropout~\cite{gal2016dropout,tsymbalov2018dropout}. Unfortunately, the reported performance of density-based scores for text generation so far has been notably low~\cite{vashurin2024benchmakring}. 

  Recent work has demonstrated that the internal states of LLMs carry a lot of information about their uncertainty~\cite{azaria-mitchell-2023-internal,chen2024eigenscore,he-etal-2024-llm,ch-wang-etal-2024-androids,vazhentsev2024unconditional}. These techniques train a supplementary model on top of the activations of LLM internal layers. However, they often rely on simplistic features and fail to incorporate more advanced, well-established density-based UQ methods, limiting their ability to capture uncertainty.

  In this work, we address this gap by adapting density-based techniques for the UQ of LLMs and
  propose a new supervised method based on density-based features. Specifically, we adapt one of the most robust methods for UQ in the classification tasks, namely Mahalanobis Distance (MD;~\citet{lee2018simple}), and train a linear regression on top of the MD scores from various layers of the LLM. These features are supplemented with a probability of the generated sequence. 
  \Cref{fig:tmd_scheme} illustrates the scheme of the proposed supervised UQ method. Our extensive experimental evaluation demonstrates that the proposed method provides substantial improvement over the state of the art.

  Our key \textbf{contributions} are as follows.
  \begin{compactitem}
    \item We conduct a comprehensive investigation of density-based UQ methods for LLMs. While previous research ~\cite{vashurin2024benchmakring} has indicated that sequence-level density-based methods are ineffective, we propose a token-level adaptation of MD that is on par with or better than state-of-the-art UQ techniques.

    \item We propose a new computationally efficient supervised method for UQ in LLMs using layer-wise density-based scores as features to improve uncertainty estimation without sacrificing the performance.

    \item We conduct a vast empirical investigation that demonstrates the effectiveness of the proposed methods for sequence-level selective classification across eleven datasets and claim-level fact-checking.
  \end{compactitem}

\section{Related Work}
  Many effective UQ methods, such as deep ensembles~\citep{NIPS2017_9ef2ed4b} and Monte Carlo (MC) dropout~\citep{gal2016dropout}, require sampling multiple predictions from a model, which leads to substantial computational and memory overheads. A key challenge in UQ is developing techniques that balance effectiveness with computational efficiency. Among the most promising approaches in this regard are density-based methods~\cite{lee2018simple,NEURIPS2020_543e8374,ddu_amersfoort,nuq_kotelevskii,yoo-etal-2022-detection}. These methods leverage latent representations of instances to model the training data distribution, then estimate how likely a new instance belongs to that distribution.~\citet{lee2018simple} propose to use Mahalanobis Distance (MD) as a measure of uncertainty for out-of-distribution detection in computer vision tasks. \citet{podolskiy2021revisiting} adapt  MD  to out-of-distribution in text classification tasks. \citet{vazhentsev-etal-2022-uncertainty,vazhentsev-etal-2023-efficient} show that it also provides high performance in selective text classification.  

  For LLMs, \citet{fomicheva-etal-2020-unsupervised} and~\citet{kuhn2023semantic} proposed UQ methods that sample multiple predictions and leverage their diversity. In the context of black-box LLMs, where we have no access to the logits or embeddings of a model, \citet{fomicheva-etal-2020-unsupervised} propose the use of lexical dissimilarity of sampled texts as a measure of uncertainty. \citet{lin2023generating} leverage a similarity matrix between responses for deeper analysis of the diversity of the sampled generations. Some methods also combine sampling diversity measures with the probability of each generation~\cite{kuhn2023semantic,duan-etal-2024-shifting,nikitin2024kernel,cheng-vlachos-2024-measuring,chen2024eigenscore,vashurin2025cocoageneralizedapproachuncertainty}.

  Recently, it was demonstrated that MD is an efficient approach for out-of-distribution detection in sequence-to-sequence models~\cite{vazhentsev-etal-2023-efficient,ren2023outofdistribution,darrin-etal-2023-rainproof}. However, for selective generation tasks, density-based methods so far have substantially underperformed compared to trivial baselines~\citep{vashurin2024benchmakring}.

  Supervised methods are another research direction for UQ of LLMs. \citet{azaria-mitchell-2023-internal} demonstrate that the internal states of the model contain information about uncertainty, and propose to train a multi-layer perceptron over the hidden LLM representation to predict the truthfulness of the model responses. \citet{he-etal-2024-llm} enhance this idea by training a deep neural network with recurrent and convolutional layers. Furthermore, this method uses embeddings from all layers and incorporates features based on the probability and the dynamics of the generated tokens through layers. In contrast to these methods, we employ a simple linear model, but focus on more accurate feature extraction from internal layers, using well-established density-based UQ methods.

\section{Background on Density-Based Methods}
  Recently, Mahalanobis distance (MD) and Robust Density Estimation (RDE) were adapted~\cite{vazhentsev-etal-2023-efficient,ren2023outofdistribution} to the text generation task by considering the marginal distribution of the training dataset. 

  Following the assumption of a Gaussian distribution of training instance representations, the MD method~\cite{lee2018simple} calculates a centroid of the training data $\mu$ and the empirical covariance matrix $\Sigma$. For a given instance $\xv$, the uncertainty score is defined as the Mahalanobis distance:
  \begin{equation*}
    U^{\text{MD}}(\xv, l) = (h_l(\xv) - \mu)^{T} \Sigma^{-1} (h_l(\xv) - \mu),
  \end{equation*}
  where $h_l(\xv)$ is a hidden representation extracted from the layer $l$.

  RDE~\cite{yoo-etal-2022-detection} operates within the reduced dimensionality of $h_l(\xv)$ via the kernel PCA decomposition. To ensure the robustness of the covariance matrix, it uses the Minimum Covariance Determinant estimate~\cite{Rousseeuw84leastmedian}. Finally, the uncertainty score is computed as the Mahalanobis distance in the reduced dimensionality.

  \citet{ren2023outofdistribution} proposed a modification of MD -- Relative Mahalanobis Distance (RMD). It takes into account a background contrastive MD score. The score aims to assess how close the test instance is to the in-domain training data compared to the background data. The uncertainty score based on RMD is given by the following equation:
  \begin{equation*}
    U^{\text{RMD}}(\xv, l) = U^{\text{MD}}(\xv, l) - U^{\text{MD}}_0(\xv, l),
  \end{equation*}
  where $U^{\text{MD}}_0(\xv, l)$ is a Mahalanobis distance computed with the centroid $\mu^0$ and the empirical covariance matrix $\Sigma^0$ calculated using the background dataset, such as C4~\cite{c4dataset}.

  For the sequence-to-sequence tasks, it was proposed to use the last encoder and decoder layer for extracting hidden representation of the model~\cite{vazhentsev-etal-2023-efficient,ren2023outofdistribution}. In contrast, recent research~\cite{azaria-mitchell-2023-internal,chen2024eigenscore} indicates that the middle layers of the model may be more suitable for decoder-only models. 

\section{Proposed Method: Token-Level Mahalanobis Distance}

  To define the method, we assume access to training data consisting of a set of prompts paired with LLM responses, each accompanied by an assessment of its correctness. The assessment can be based on ground truth answers (as in tasks like question-answering, machine translation, or summarization) or through alternative means, such as human annotation or another big LLM.

\subsection{Layer-Wise Uncertainty Score}

\paragraph{Embedding extraction.}
  First of all, we need to extract embeddings of instances in the training dataset. We note that previous works use sequence-level embeddings, which are essentially an average of token-level embeddings. Recent studies~\cite{azaria-mitchell-2023-internal,chen2024eigenscore} note that sequence embeddings might be useless for UQ with LLMs and propose to use embeddings of the last or the first generated token, as they encode useful information for estimating the truthfulness of the entire generation. We acknowledge that this property may not always hold, as the informative tokens are likely to vary depending on the specific task. In our method, we first compute individual token-level uncertainty scores and then aggregate them into a sequence-level score. 

\paragraph{Embedding selection.}
  To construct a covariance matrix and centroid for MD, a model training set is required. 
  However, unlike standard text classification tasks, where training sets are typically limited and accessible during the development of an ML-based application, the pre-training data for general-purpose LLMs is extremely large and usually not publicly available. Moreover, even if this data were available, LLM performance on it would likely be not homogeneous and could be low for certain tasks. Therefore, to construct the parameters for MD, we propose selecting a subset of token embeddings from high-quality LLM responses. 

  From the responses generated in the training set, we select a subset of token embeddings that correspond to responses that meet a defined correctness criterion. Let $\TC$ be a training set of input prompts and $|\TC| = N_{|\TC|}$. For each prompt $\xv^j \in \TC$, the model generates a response as a sequence $\tilde{\yv}^j = \tv^j_1, \dots, \tv^j_{N_j}$, where $N_j$ is a length of the $j$-th generation and $\tv^j_i, i\in[1, \dots, N_j]$ is an $i$-th token in the response. We define a set of selected tokens as $\DC = \{\tv^j_i\colon\QC(\tilde{\yv}^j) > \tau, i\in[1, \dots, N_j], j\in[1, \dots, N_{|\TC|}] \}$, where $\QC(\cdot)$ is a quality metric and $\tau$ is a given threshold. Then $\EC_l = \{h_l(\tv)\colon\tv\in\DC\}$ is the set of selected token embeddings. The correctness criterion helps filter out low-quality responses. Depending on the dataset in the experiment, exact match and AlignScore are employed as quality metrics. The correctness criterion used for token selection is described in \Cref{sec:setup}. 

\paragraph{Layer-wise scores.}
  For each layer $l=1, \dots, L$ of the model, we compute the covariance matrix $\Sigma_{\EC_l}$ and the centroid $\mu_{\EC_l}$ using the set of selected token embeddings $\EC_l$. For each token from the generated sequence $\tilde{\yv}^k = \tv^k_1, \dots, \tv^k_{N_k}$, we compute the layer-wise MD as follows:
  \begin{equation*}
    U^{\text{MD}}(\tv^k_i, l) \!=\! \bigl(h_l(\tv^k_i) - \mu_{\EC_l}\bigr)^{T} \Sigma_{\EC_l}^{-1} \bigl(h_l(\tv^k_i) - \mu_{\EC_l}\bigr).
  \end{equation*}
  For the token-level RMD, we additionally compute the background covariance matrix $\Sigma^0_l$ and the background centroid $\mu^0_l$ using the embeddings of all generated tokens for the input prompts from some background dataset. 

  Finally, the uncertainty score of the entire generated sequence $\tilde{\yv}^k$ is the \textit{Average Token-level Mahalanobis Distances (ATMD)} over $\tv^k_i, i=1, \dots, N_k$ (for RMD, we designate it as ATRMD).

\subsection{Linear Regression on Layer-Wise Scores}
\label{sec:supervised}
  The ATMD and ATRMD scores can be computed on various layers. \citet{azaria-mitchell-2023-internal} indicate that the best-performing layer may vary depending on the generation task. To effectively integrate information from multiple layers, we propose training a regression model on top of the layer-wise scores.

  For a generation $\tilde{\yv}^k$, we construct a vector of features based on $\text{ATMD}$ or $\text{ATRMD}$: $f^{*}(\tilde{\yv}^k)=[U^{*}(\tilde{\yv}^k, 1), \dots, U^{*}(\tilde{\yv}^k, L)]$ (we use $*$ instead of $\text{ATMD}$ or $\text{ATRMD}$). To learn the uncertainty of the generation, we define target variables as negative values of a quality metric for generations $\tilde{\yv}^k$: $\qv^k= - \QC(\tilde{\yv}^k)$. We note that the features $f^{*}(\tilde{\yv}^k)$ might be highly correlated with each other (a multicollinearity problem;~\citet{multicollinearity}), which makes linear models to overfit~\cite{mitigating_multicollinearity}. To make our features more robust, we use top $N=10$ components from the PCA decomposition of feature vectors: $\tilde{f^{*}}(\tilde{\yv}^k) = \text{PCA}_N \bigl(f^{*}(\tilde{\yv}^k)\bigr)$.

  We train the machine learning model $G(\cdot)$ to predict an uncertainty score as follows:
  \begin{enumerate}[topsep=0pt,itemsep=-1ex,partopsep=1ex,parsep=1ex]
    \item Split the entire training dataset $\TC$ into two parts $\TC_1$ and $\TC_2$.
    \item Using $\TC_1$, construct $\EC_l, l \in [1, \dots, L]$ and fit layer-wise covariance matrices and centroids. ATRMD also fits layer-wise background covariance matrices and background centroids.
    \item For each generation $\tilde{\yv}^k, k = 1, \dots, |\TC_2|$ for the prompts from $\TC_2$, compute features $\tilde{f^{*}}(\tilde{\yv}^k)$ and targets $\qv^k$.
    \item Train the machine learning model $G^{*}(\cdot)$ to predict the targets $\qv^k$ using the features $\tilde{f^{*}}(\tilde{\yv}^k)$, $k = 1, \dots, |\TC_2|$. In our work, we use linear regression models as $G^{*}(\cdot)$.
    \item Re-estimate layer-wise parameters of the distribution using the entire training dataset $\TC$.
  \end{enumerate}

  Finally, the supervised uncertainty score for a test generation $\tilde{\yv}^k$ based on token-level MD or RMD, namely SATMD or SATRMD is:
  \begin{equation*}
    U^{\text{S*}}(\tilde{\yv}^k) = G^{*}\bigl(\tilde{f^{*}}(\tilde{\yv}^k)\bigr).
  \end{equation*}

  Following~\citet{he-etal-2024-llm}, we also experiment with adding the sequence probability $P(\tilde{\yv}^k \mid \xv^k)$ as an additional feature to the features vector: $\tilde{f^{*}}_{prob}(\tilde{\yv}^k) = [\tilde{f^{*}}(\tilde{\yv}^k); P(\tilde{\yv}^k \mid \xv^k)]$, and get
  \begin{equation*}
    U^{\text{S*+MSP}}(\tilde{\yv}^k) = G^{*}\bigl(\tilde{f^{*}}_{prob}(\tilde{\yv}^k)\bigr).
  \end{equation*}

\subsection{Hybrid Score}
  In addition, we explore Hybrid Uncertainty Quantification (HUQ;~\citet{vazhentsev-etal-2023-hybrid}), which empirically combines multiple uncertainty scores. Using HUQ, we combine sequence probability $U_{\text{1}}(\tilde{\yv}^k) = 1 - P(\tilde{\yv}^k \mid \xv^k)$ and the proposed SATMD or SATRMD scores: $U_{\text{2}}(\tilde{\yv}^k) = U^{\text{S*}}(\tilde{\yv}^k)$. The hyperparameters of HUQ are tuned on the $\TC_2$ dataset. A detailed description of the HUQ method is given in~\Cref{sec:huq}.

\section{Experiments}

  \begin{figure*}[t!]
    \centering
    \includegraphics[trim={0.cm 0.cm 0.cm 0.cm},clip,width=1.\linewidth]
    {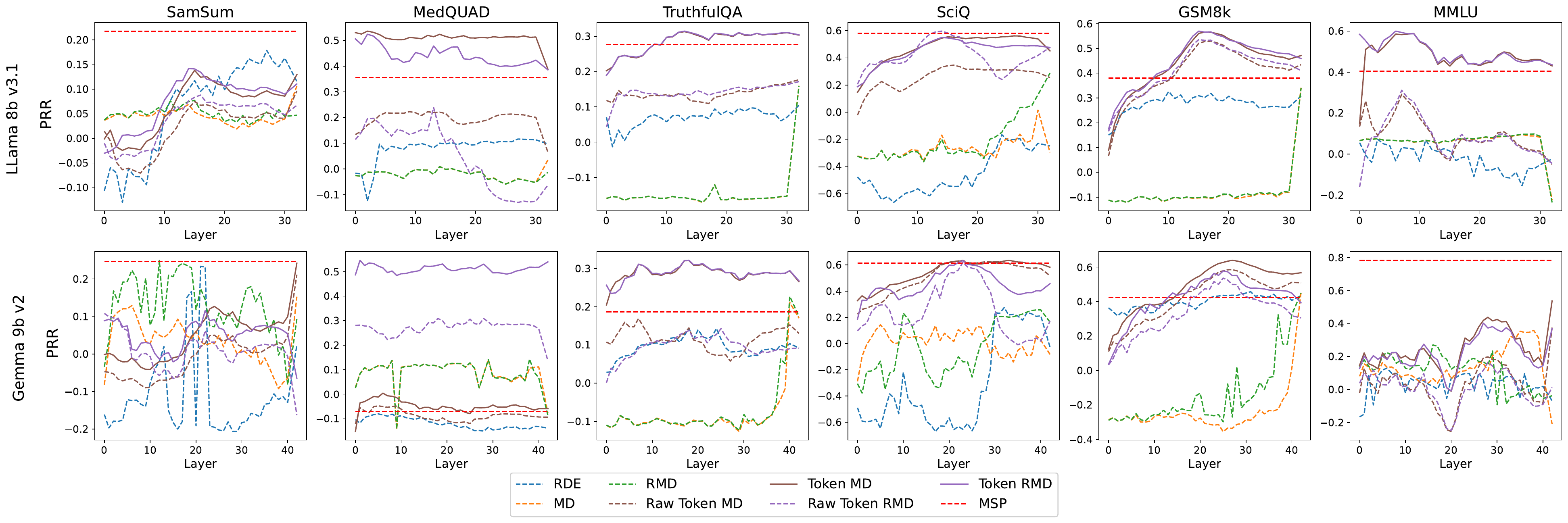}
    \caption{
    Performance of embeddings from various layers in density-based scores. PRR$\uparrow$ for density-based methods computed using embeddings from various layers of Llama 8b v3.1 (upper row) and Gemma 9b v2 (lower row) models. Raw ATMD/ATRMD denotes a corresponding method without selecting embeddings using the correctness criterion. Higher values indicate better results.
    }
    \label{fig:tmd_layerwise}
  \end{figure*}

\subsection{Experimental Setup}
\label{sec:setup}
  For the experimental evaluation, we employ the LM-Polygraph framework~\cite{fadeeva2023lm,vashurin2024benchmakring}. We consider two tasks: (1) sequence-level selective generation~\cite{ren2023outofdistribution}, in which we can ``reject'' untruthful generations based on provided uncertainty; (2) claim-level fact-checking~\cite{fadeeva2024factchecking}, where we aim to identify nonfactual claims in long generations, consisting of several claims. 

\paragraph{Metrics.}
  To evaluate the quality of UQ methods on the selective generation task, we use the standard Prediction Rejection Ratio (PPR) metric~\cite{malinin21uncertainty,vashurin2024benchmakring}. This metric measures the correctness of the ranking of generations based on uncertainty relative to a specified quality metric. PRR computes the area under the Prediction Rejection (PR) curve, which is constructed by sequentially rejecting the most uncertain generation and calculating the average quality for all stored generations at each possible threshold. Subsequently, this area is normalized by scaling between the PR curve for the random selection and oracle selection. A higher value of the PRR corresponds to a better quality of selective generation. Following previous work~\cite{vashurin2024benchmakring}, we use ROUGE-L, Accuracy, and AlignScore~\cite{zha-etal-2023-alignscore} as text generation quality metrics.

  For claim-level fact-checking, we follow previous work~\cite{fadeeva2024factchecking} and consider this task as a binary classification problem. We utilize the ROC-AUC and PR-AUC metrics, where nonfactual claims represent a positive class.

\paragraph{Models.}
  For selective generation, we experiment with two state-of-the-art LLMs in their size: Llama@8b~v3.1~\cite{llama3} and Gemma~9b~v2~\cite{gemma2}. For fact-checking, we utilize Mistral 7b v0.1 Instruct~\cite{mistral}. The inference hyperparameters are presented in \Cref{tab:llm_hyperparameters} in \Cref{sec:hyperparameters}.

\paragraph{Datasets.}

  \input{./llama_aggregations}

  We consider several text generation tasks, including text summarization (TS), question-answering (QA) with long free-form answers, QA based on reading comprehension, QA with short free-form answers, and multiple-choice QA. Dataset statistics are presented in~\Cref{tab:dataset_stat} in~\Cref{sec:datasets}. For TS, we utilize XSum~\cite{xsum}, SamSum~\cite{gliwa-etal-2019-samsum}, and the CNN/DailyMail~\cite{see-etal-2017-get} dataset. For QA with long free-form answers, we use PubMedQA~\cite{jin-etal-2019-pubmedqa}, MedQUAD~\cite{medquad}, TruthfulQA~\cite{lin-etal-2022-truthfulqa}, and GSM8k~\cite{gsm8k}. For reading comprehension, we use CoQA~\cite{coqa} and SciQ~\cite{welbl-etal-2017-crowdsourcing}. For QA with short free-form answers, we use TriviaQA~\cite{joshi-etal-2017-triviaqa}. The last three datasets represent the common benchmarks for evaluating UQ methods in previous work~\cite{kuhn2023semantic,duan-etal-2024-shifting,lin2023generating}. For multiple-choice QA, we utilize MMLU~\cite{mmlu}, which is a common dataset for evaluating LLMs.

\paragraph{UQ baselines.}
  In an experimental evaluation, we compare the proposed methods against several UQ baselines, including trivial yet robust information-based methods such as Maximum Sequence Probability (MSP) and Perplexity~\cite{fomicheva-etal-2020-unsupervised}, and consistency-based methods considered state-of-the-art for LLMs~\cite{vashurin2024benchmakring}: Lexical Similarity based on ROUGE-L~\cite{fomicheva-etal-2020-unsupervised}, black-box methods (DegMat, Eccentricity, EigValLaplacian; \citet{lin2023generating}), Semantic Entropy~\cite{kuhn2023semantic}, and Shifting Attention to Relevance (SAR; \citet{duan-etal-2024-shifting}). Furthermore, to ensure the robustness of the proposed methods, the suite of baselines in our experiments also includes methods that utilize model internal states: Factoscope~\cite{he-etal-2024-llm}, SAPLMA~\cite{azaria-mitchell-2023-internal}, and EigenScore~\cite{chen2024eigenscore}. The first two are supervised methods, while EigenScore is unsupervised. Following the previous works~\cite{azaria-mitchell-2023-internal,chen2024eigenscore}, we use embeddings from the middle layer of the model for the latter two methods. For the methods that require sampling, we sample five generations for each input text.

\paragraph{Configuration of ATMD/ATRMD.}
  In in-domain experimental evaluation on the SciQ, CoQA, TriviaQA, MMLU, and GSM8k datasets, we select token embeddings used to construct the covariance matrix and centroids for ATMD and ATRMD from generations that are fully accurate according to the exact match criterion. On other datasets, we utilize generations with AlignScore greater than 0.3. Raw ATMD/ATRMD denotes a corresponding method without selecting embeddings using the correctness criterion.

\subsection{Results}
\paragraph{Layer-wise comparison of density-based methods.}
\label{sec:layer_wise_results}
  \Cref{fig:tmd_layerwise} presents the layer-wise comparison of various sequence-level density-based approaches for selective generation for the Llama 8b v3.1 and Gemma 9b v2 models. These results demonstrate the presence of robust patterns across the majority of datasets and models. 
  
  Consistent with the findings of~\citet{vashurin2024benchmakring}, we observe that in most cases, density-based methods that use sequence-level embeddings (MD, RMD, and RDE) yield PRR scores that are close to or below zero, indicating performance comparable to random selection. Only for GSM8k, these methods provide meaningful uncertainty scores, but they still do not outperform the basic MSP baseline. Furthermore, we see that using sequence-level embeddings derived from internal layers does not improve the performance of density-based methods; they usually perform better when using embeddings from the last layer. 
  \input{./llama}

  MD that uses token-level embeddings performs consistently better than the MD based on sequence-level embeddings for all datasets except SamSum, where all methods perform similarly to each other.
  Moreover, density-based methods that compute MD using token-level embeddings from internal layers outperform those that rely on embeddings from the top layers. While for SamSum and MMLU with the Gemma 9b v2 model, ATMD achieves the best performance using the last layer embeddings, for all other cases the best performance is achieved by using embeddings from the middle layers. This finding is consistent with previous research~\cite{azaria-mitchell-2023-internal,chen2024eigenscore}. 
  
  Using the selection of correct generations from the training dataset for fitting the covariance matrix and centroid is key to achieving good performance of the methods based on token-level MDs. ATMD/ATRMD consistently outperform raw ATMD/ATRMD. The highest difference was observed on the MedQUAD and TruthfulQA datasets, where using selection improved PRR by 0.2-0.3. 
  
\paragraph{Comparison of sequence-level aggregations.}
  \Cref{tab:llama_agg_results,tab:gemma_agg_results} in \Cref{sec:aggregation} present the comparison of various sequence-level supervised approaches for selective generation for the Llama 8b v3.1 and Gemma 9b v2 models. The results demonstrate that SATMD and SATRMD provide stable and robust performance, which is often superior or equal to the performance of the MD/RMD using the embeddings from the best layer. As anticipated, the incorporation of MSP as an additional feature or combining it using HUQ significantly improved the performance of SATMD/SATRMD. While on average by mean rank, the best performance across various datasets was achieved by SATMD+MSP, for XSum and SciQ, HUQ-SATRMD exhibited a slight improvement. It is also noteworthy that using RMD led to a consistent improvement in performance compared to the original MD for all variants of the supervised method. 

\paragraph{Main results on the selective generation tasks.}
  The main results on the selective generation tasks for the Llama 8b v3.1 and Gemma 9b v2 models are presented in \Cref{tab:llama_results,tab:gemma_results} in \Cref{sec:overall_results}. In the summarization task, our supervised methods outperform all the baselines on XSum and SamSum.
  HUQ-SATRMD achieves the best performance on the XSum and SamSum datasets in terms of PRR-ROUGE-L. For SamSum, the SATRMD+MSP method significantly outperforms other methods in terms of PRR-AlignScore. For the CNN dataset, the proposed methods demonstrate the second-best results in terms of PRR-ROUGE-L, but they substantially fall behind unsupervised UQ techniques in terms of PRR-AlignScore.

  In the QA tasks with long answers (PubMedQA, MedQUAD, TruthfulQA, and GSM8k), SATRMD+MSP consistently demonstrates the best performance, with a notable margin over best supervised and unsupervised techniques, while HUQ-SATRMD ranks second. In the reading comprehension task, HUQ-SATRMD is the best-performing method. Meanwhile, on the MMLU dataset, SATRMD+MSP is the most effective method, significantly outperforming HUQ-SATRMD and other state-of-the-art baselines.

  Considering QA with short answers on CoQA, we observe that HUQ-SATRMD performs on par with the MSP baseline, while on the SciQ dataset performs the best with a large margin. On TriviaQA, SATRMD+MSP outperforms the MSP baseline, underperforming only sampling-based methods that require much more computation time. 

  Overall, we can conclude that HUQ-SATRMD is the most effective method for summarization and reading comprehension tasks, where it significantly outperforms state-of-the-art unsupervised UQ methods. For all other QA datasets, including those with long answers and tasks requiring internal knowledge, the best performance is demonstrated by SATRMD+MSP.

\paragraph{Out-of-domain generalization.}
  \input{./generalization_llama_v2}

    \begin{figure*}[t!]
    \centering
    \includegraphics[trim={0.cm 0.cm 0.cm 0.cm},clip,width=1.\linewidth]{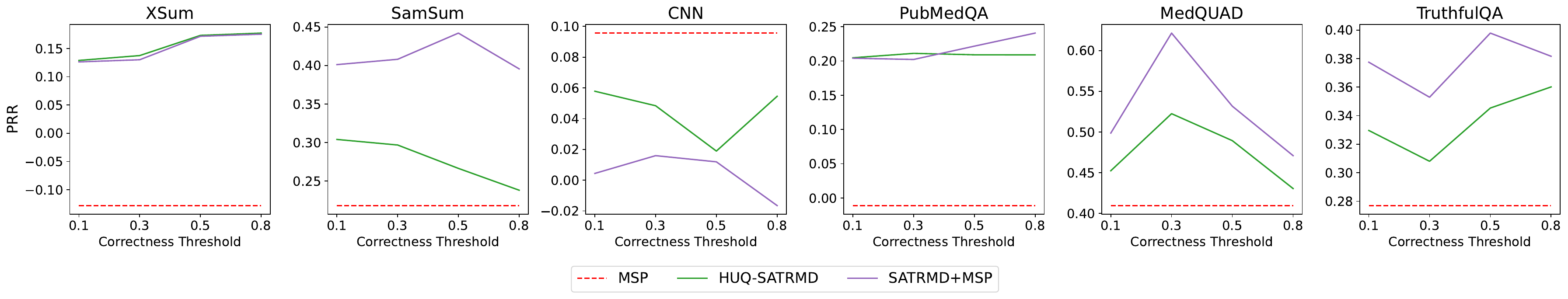}
    \caption{
    Dependency of PRR$\uparrow$ of the SATRMD+MSP and HUQ-SATRMD methods on the correctness threshold for the embedding selection for the centroid and covariance matrix for MD for the Llama 8b v3.1 model. Higher values indicate better results.
    }
    \label{fig:tmd_threshold}
  \end{figure*}

  \Cref{tab:llama_gen_results} presents a comparison of various sequence-level methods for the selective generation task for the Llama 8b v3.1 model in the out-of-domain settings. We train and evaluate supervised methods using a leave-one-out approach: train on all datasets except one, which is left for testing.
  For each evaluation dataset, the training set is composed of 400 instances sampled from each of the remaining datasets. We use the negative AlignScore generation quality metric as a target for all considered datasets in this setting. 

  The performance of supervised methods in the out-of-domain setting shows a significant decline compared to the in-domain setting. Despite this, HUQ-SATRMD achieves the best results according to the mean rank, outperforming unsupervised state-of-the-art methods across the majority of datasets and metrics, except MSP on SamSum and MedQUAD in terms of PRR-ROUGE-L. Notably, when testing on the MMLU dataset, the training data consists of texts from summarization tasks and free-form QA, which differ significantly from the multiple-choice QA format used in MMLU. Nevertheless, the strong performance on MMLU demonstrates the potential of our supervised method HUQ-SATRMD for broad generalization.
  
  Other supervised methods, including SATRMD+MSP and the baselines, show significantly poorer results in the out-of-domain setting. SATRMD+MSP underperforms MSP on several datasets, including SamSum, MedQUAD, and TruthfulQA. SAPLMA and Factoscope are not able to provide meaningful uncertainty scores, lagging significantly behind unsupervised UQ methods.

\paragraph{Fact-checking results.}

\input{./factchecking}

  \Cref{tab:factcheking_results} presents a comparison of various claim-level methods for the fact-checking task using the Mistral 7b v0.1 Instruct model. The baseline supervised method SAPLMA performs similarly to a random choice. Our method SATRMD provides meaningful uncertainty scores slightly outperforming Maximum Claim Probability (MCP). We note that CCP, like MCP, is also based on the probabilities derived from the model output but demonstrates better performance than MCP. Consequently, we combine CCP with SATRMD to provide more effective claim-level fact-checking. The results demonstrate that HUQ-SATRMD achieves the best results in terms of ROC-AUC, outperforming CCP by 3.4\%, while in terms of PR-AUC, SATRMD+CPP is the best, outperforming CCP by 2.6\%. These results demonstrate that the proposed SATRMD methods are effective not only for sequence-level uncertainty quantification but also for estimating uncertainty on the claim level.

  \begin{figure*}[t!]
    \centering
    \includegraphics[trim={0.cm 0.cm 0.cm 0.cm},clip,width=1.\linewidth]
    {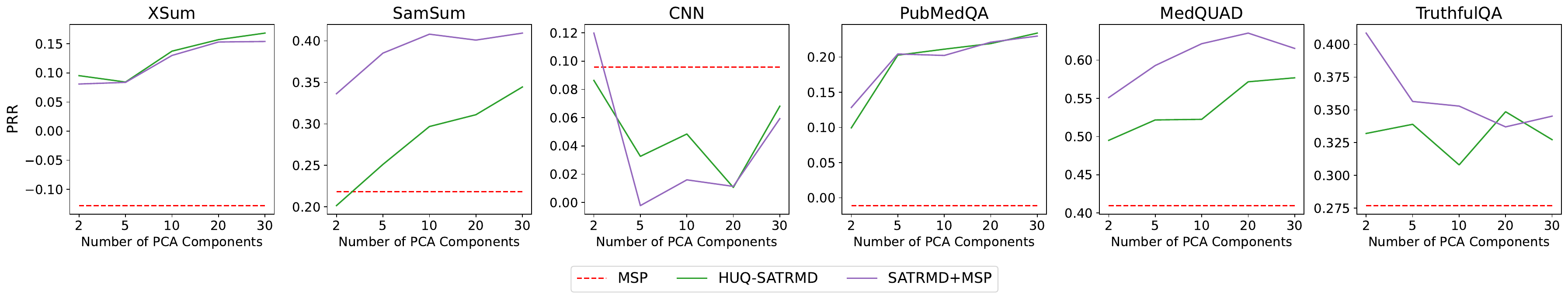}
    \caption{
    Dependency of PRR$\uparrow$ of the SATRMD+MSP and HUQ-SATRMD methods on the number of the PCA components for the features of linear regression for the Llama 8b v3.1 model. Higher values indicate better results.
    }
    \label{fig:tmd_pca}
  \end{figure*}

\paragraph{Impact of training data size.}
  \Cref{fig:tmd_trainsize} in \Cref{sec:trainsize_results} illustrates the dependency of the performance of supervised UQ methods on the size of the training data. As expected, the optimal results on all datasets are achieved when the maximum number of training instances is used. 
  Nevertheless, for all datasets, except SamSum and MedQUAD, the results obtained with 200-500 training instances are only slightly lower than with 2,000-5,000 instances. Furthermore, even with fewer than 200 training instances for MedQUAD, GSM8k, and MMLU, HUQ and SATRMD+MSP are able to substantially outperform the MSP method. These results demonstrate the robustness of the proposed methods with respect to the size of the training dataset.

\paragraph{Impact of the correctness threshold.}
  \Cref{fig:tmd_threshold} presents the dependence of the performance of the SATRMD+MSP and HUQ-SATRMD methods on the correctness threshold used for the embedding selection for computing the centroid and covariance matrix for MD. 

  The results demonstrate that the proposed methods are generally not sensitive to the correctness threshold and consistently show high performance. However, for the MedQUAD dataset, we can see the results with a threshold of 0.3 are significantly better than those with other thresholds. Specifically, lower thresholds (e.g., 0.1) result in selecting the embeddings corresponding to incorrect instances, while higher thresholds (e.g., 0.8) exclude some embeddings associated with correct instances. Both scenarios result in suboptimal estimation of the centroid and covariance matrix, leading to a slight degradation in overall performance.

\paragraph{Impact of the number of PCA components.}
  \Cref{fig:tmd_pca} illustrates the impact of the number of PCA components used for the features of linear regression on the performance of SATRMD+MSP and HUQ-SATRMD methods. The best performance is achieved with 10 or 20 components for most datasets. Only for CNN and TruthfulQA, using just 2 components yields slightly better results than using more. Overall, these results indicate that our choice of 10 components is well-balanced on average across multiple datasets. We also observe that results with more than five PCA components remain stable across all datasets, showing minimal variation. Therefore, methods based on RMD are not sensitive to the precise choice of the number of PCA components.

\paragraph{Computational efficiency.}
  \Cref{tab:comp_efficiency} in~\Cref{sec:comp_eff} summarizes the average runtime per instance for each UQ method, along with the percentage overhead compared to standard LLM inference. State-of-the-art UQ methods that require sampling from the LLM multiple times (Semantic Entropy, SAR, Lexical Similarity) introduce a huge computational overhead (315-700\%). In contrast, the proposed methods HUQ-SATRMD and SATRMD+MSP introduce minimal overhead (5.3-7.6\%), which makes them much more practical.


\section{Conclusion}
  We have introduced a series of new supervised UQ methods based on layer-wise features derived from the Mahalanobis distance. We show that calculating MD over token-level embeddings yields much better results than previous attempts that leverage sequence-level embeddings. Training a linear regression on top of the layer-wise scores allows us to produce even better uncertainty scores and outperform the state-of-the-art supervised and unsupervised UQ methods in selective classification across eleven datasets and in claim-level fact-checking.
  We also show that the proposed methods are computationally efficient and have the potential for generalization, which makes them useful in real-world LLM-based applications.

  In future work, we aim to improve the generalization capabilities of the supervised UQ methods on out-of-domain data by investigating new features and a more robust training pipeline.

\section*{Limitations}
  Our approach is supervised, which means that its performance depends on the quality and size of the data available for supervision. We evaluated the robustness of the approach to dataset variation, which demonstrates that the method does not significantly degrade its quality compared to the target dataset. Nevertheless, we observe certain performance drops; thus, the resulting UQ method should be used with care beyond the supervision domain.

  We did not test our method on very large LLMs, such as LLaMA 3 70b, as we were limited to using 7-9b models due to constraints in our computational resources.

\section*{Ethical Considerations}
  In our study, we focused on open-source LLMs and datasets that are not designed to produce harmful content. However, LLMs can still generate potentially harmful texts that may impact various groups. Uncertainty quantification techniques offer a way to enhance the reliability of neural networks and can even be used to detect harmful outputs, though this is not our focus.

  Although our proposed method shows substantial performance improvements, it may sometimes incorrectly flag safe and accurate generated text as having high uncertainty. While we explicitly benchmarked the method on robustness to the task change, its applicability across various tasks remains limited.

\section*{Acknowledgments}
We thank the anonymous reviewers for their valuable suggestions, which significantly contributed to improving this paper. The work of Alexander Panchenko was partially supported by the joint MTS-Skoltech laboratory.

\bibliography{custom}

\appendix
\onecolumn

\section{Additional Experimental Results}
\label{sec:experiments}

\subsection{Comparison of Sequence-Level Aggregations}
\label{sec:aggregation}
  \input{./gemma_aggregations}

\subsection{Selective Generation Results}
\label{sec:overall_results}
  \input{./gemma}

\subsection{Dependency on the Size of the Training Dataset}
\label{sec:trainsize_results}
\begin{figure*}[ht!]
    \centering
    \includegraphics[trim={0.cm 0.cm 0.cm 0.cm},clip,width=1.\linewidth]{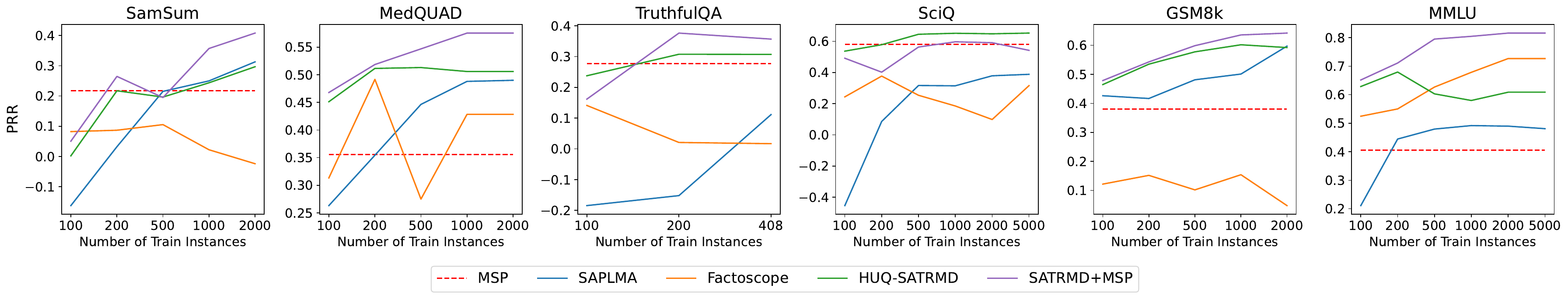}
    \caption{
    Dependency of PRR$\uparrow$ of the supervised methods on the size of the training dataset for the Llama 8b v3.1 model. Higher values indicate better results.
    }
    \label{fig:tmd_trainsize}
  \end{figure*}

  \Cref{fig:tmd_trainsize} presents the results when varying the size of the training dataset for the supervised methods. We train the linear regression model on the training datasets of size: 100, 200, 500, 1000, 2000, and additionally on a training dataset of 5000 instances for SciQ and MMLU. Since the TruthfulQA dataset consists of only 817 instances, of which we use 409 instances as the test subset, we train linear regression on the training datasets of sizes: 100, 200, and 408. 

\section{Hybrid Uncertainty Quantification}
\label{sec:huq}
  We combine the sequence probability $U_{\text{1}}(\tilde{\yv}^k) = 1 - P(\tilde{\yv}^k \mid \xv^k)$ with the SATMD and SATRMD methods $U_{\text{2}}(\tilde{\yv}^k) = U^{\text{S*}}(\tilde{\yv}^k)$. For a given $\TC_1$ and $\TC_2$ from~\Cref{sec:supervised}, and trained SATRMD method, we fit HUQ hyperparameters on the $\TC_2$ set.

  Following~\citet{vazhentsev-etal-2023-hybrid}, we define the set of in-distribution instances from $\TC_2$ as follows: $\TC_{\text{ID}} = \{\xv \in \TC_2\colon U_{\text{2}}(\xv) \leq \delta_{\min}\}$. We define the set of arbitrary in-distribution instances $\XSet_{\text{ID}} = \{\xv\colon U_{\text{2}}(\xv) \leq \delta_{\min}\}$ and ambiguous in-distribution instances $\XSet_{\text{IDA}} = \{\xv {\in \XSet_{\text{ID}}}\colon U_{\text{1}}(\xv) > \delta_{\max}\}$ using $\delta_{\min}$, $\delta_{\max}$ are thresholds selected on the $\TC_2$ dataset. 

  To make different uncertainty scores comparable, we define a ranking function $R(\uv, \mathfrak{D})$ as a rank of $\uv$ over a sorted dataset $\mathfrak{D}$, where $\uv_1 > \uv_2$ implies $R(\uv_1, \mathfrak{D}) > R(\uv_2, \mathfrak{D})$. We compute the total uncertainty $U_{\text{T}}(\xv)$ as a linear combination $U_{\text{T}}(\xv) = \!\!(1 - \alpha) R(U_{\text{2}}(\xv), \TC_2) + \alpha R(U_{\text{1}}(\xv), \TC_2)$, where $\alpha$ is a hyperparameter selected on the $\TC_2$ dataset. As a result, we define HUQ as follows:
  \begin{equation*}
    U_{\text{HUQ}}(\xv)
    = 
    \begin{cases}
      R(U_{\text{1}}(\xv), \TC_{\text{ID}}), \forall \xv \in \XSet_{\text{ID}} \setminus \XSet_{\text{AID}}, \\
      R(U_{\text{1}}(\xv), \TC_2),  \forall \xv \in \XSet_{\text{AID}}, \\
      U_{\text{T}}(\xv), \forall \xv \notin \XSet_{\text{ID}}.
    \end{cases}
  \end{equation*}

\section{Computational Efficiency}
\label{sec:comp_eff}
  \input{./comp_time}

\section{Computational Resources}
  All experiments were conducted on a cluster with 6 NVIDIA H100 GPUs. The total time for all conducted experiments for all models across all datasets is approximately 400 GPU hours.

\section{Dataset Statistics}
\label{sec:datasets}
  \input{./datasets}

\clearpage
\section{Inference Hyperparameters}
\label{sec:hyperparameters}
  \input{./llms_hp}

\end{document}

%% file: llama_aggregations.tex
\begin{table*}[!ht] \resizebox{\textwidth}{!}{\begin{tabular}{l||cc|cc|cc|cc|cc|c|c|c|c|c|c|c}
\toprule
\multirow{2}{*}{\textbf{UQ Method}} & \multicolumn{2}{c|}{\textbf{XSUM}} & \multicolumn{2}{c|}{\textbf{SamSum}} & \multicolumn{2}{c|}{\textbf{CNN}}& \multicolumn{2}{c|}{\textbf{PubMedQA}} & \multicolumn{2}{c|}{\textbf{MedQUAD}} & \multicolumn{1}{c|}{\textbf{TruthfulQA}}& \multicolumn{1}{c|}{\textbf{CoQA}} & \multicolumn{1}{c|}{\textbf{SciQ}} & \multicolumn{1}{c|}{\textbf{TriviaQA}} & \multicolumn{1}{c|}{\textbf{GSM8k}} & \multicolumn{1}{c|}{\textbf{MMLU}} & \multirow{2}{*}{\textbf{Mean Rank}} \\ \cline{2-17}
    & \textbf{ROUGE-L} & \textbf{AlignScore} & \textbf{ROUGE-L} & \textbf{AlignScore} & \textbf{ROUGE-L} & \textbf{AlignScore} & \textbf{ROUGE-L} & \textbf{AlignScore} & \textbf{ROUGE-L} & \textbf{AlignScore} & \textbf{AlignScore} & \textbf{AlignScore} & \textbf{AlignScore}& \textbf{AlignScore} & \textbf{Accuracy}& \textbf{Accuracy}& \\
\midrule
SATMD & \large\cellcolor[rgb]{0.9836582578333333,0.8287354144039216,0.7690800753647059} .189 & \large\cellcolor[rgb]{0.9727701494549019,0.8993028702666667,0.8615527086490196} .062 & \large\cellcolor[rgb]{0.7473192431058824,0.8165111307921569,0.9894914084686275} .369 & \large\cellcolor[rgb]{0.61490285,0.649358983,0.8768415764999999} .232 & \large\cellcolor[rgb]{0.6426363887647059,0.690064711317647,0.9117342756235294} .059 & \large\cellcolor[rgb]{0.8980319349294118,0.9243466028941176,0.9667357341529412} .030 & \large\cellcolor[rgb]{0.9090282467058823,0.927794838772549,0.9573188082745099} .217 & \large\cellcolor[rgb]{0.9824176791176471,0.8723068372941176,0.8216194376764706} .183 & \large\cellcolor[rgb]{0.9671527274529412,0.762958755827451,0.7061432370215687} \underline{.471} & \large\cellcolor[rgb]{0.9683898066058824,0.766374750054902,0.7090466697431372} .552 & \large\cellcolor[rgb]{0.61490285,0.649358983,0.8768415764999999} .230 & \large\cellcolor[rgb]{0.61490285,0.649358983,0.8768415764999999} .310 & \large\cellcolor[rgb]{0.8980319349294118,0.9243466028941176,0.9667357341529412} .385 & \large\cellcolor[rgb]{0.61490285,0.649358983,0.8768415764999999} .293 & \large\cellcolor[rgb]{0.9112102024705883,0.9284487582705883,0.9553975652941177} .587 & \large\cellcolor[rgb]{0.8492270432274509,0.8997249420568627,0.9922887283509805} .623 & \large\cellcolor[rgb]{0.61490285,0.649358983,0.8768415764999999} 4.75 \\
SATRMD & \large\cellcolor[rgb]{0.8587172724588235,0.5255587742117647,0.5793683038509804} \underline{.389} & \large\cellcolor[rgb]{0.864597965917647,0.5433394684235294,0.5836202017019608} \underline{.181} & \large\cellcolor[rgb]{0.61490285,0.649358983,0.8768415764999999} .338 & \large\cellcolor[rgb]{0.7880256462666666,0.8543899393333334,0.9988772960000001} .282 & \large\cellcolor[rgb]{0.852836579,0.50777808,0.575116406} \textbf{.138} & \large\cellcolor[rgb]{0.61490285,0.649358983,0.8768415764999999} -.004 & \large\cellcolor[rgb]{0.9028614815235294,0.6299065681294118,0.6152808287} .351 & \large\cellcolor[rgb]{0.852836579,0.50777808,0.575116406} \textbf{.213} & \large\cellcolor[rgb]{0.9847608008647059,0.8504164338117647,0.7937540087647059} .459 & \large\cellcolor[rgb]{0.9441952453705882,0.708851458745098,0.6639489555019608} \underline{.559} & \large\cellcolor[rgb]{0.7152534441254902,0.7824413707294118,0.974444709590196} .251 & \large\cellcolor[rgb]{0.6667452396901961,0.7231326097019608,0.9372260848549019} .324 & \large\cellcolor[rgb]{0.9747269947588235,0.7861939559392157,0.7267214884509805} .542 & \large\cellcolor[rgb]{0.7717200969960785,0.8400012947058824,0.9965252584941177} .394 & \large\cellcolor[rgb]{0.9824176791176471,0.8723068372941176,0.8216194376764706} .606 & \large\cellcolor[rgb]{0.61490285,0.649358983,0.8768415764999999} .504 & \large\cellcolor[rgb]{0.9176723556764705,0.9302569986470588,0.9494852049705882} 3.69 \\\midrule
HUQ-SATMD & \large\cellcolor[rgb]{0.61490285,0.649358983,0.8768415764999999} -.326 & \large\cellcolor[rgb]{0.61490285,0.649358983,0.8768415764999999} -.123 & \large\cellcolor[rgb]{0.9844470791666666,0.8397397817137255,0.7814061455764706} \underline{.441} & \large\cellcolor[rgb]{0.6691882557215687,0.7264093044156863,0.9396585384392157} .249 & \large\cellcolor[rgb]{0.61490285,0.649358983,0.8768415764999999} .055 & \large\cellcolor[rgb]{0.8980319349294118,0.9243466028941176,0.9667357341529412} .030 & \large\cellcolor[rgb]{0.61490285,0.649358983,0.8768415764999999} .088 & \large\cellcolor[rgb]{0.61490285,0.649358983,0.8768415764999999} .128 & \large\cellcolor[rgb]{0.8042736801705882,0.8678626149117648,0.9996769126490196} .417 & \large\cellcolor[rgb]{0.61490285,0.649358983,0.8768415764999999} .465 & \large\cellcolor[rgb]{0.9068462909411765,0.9271409192745098,0.959240051254902} .289 & \large\cellcolor[rgb]{0.9218513348607843,0.6650340647294117,0.635032405109804} \underline{.450} & \large\cellcolor[rgb]{0.9497671903627452,0.7203459010784313,0.6720534316176471} \underline{.577} & \large\cellcolor[rgb]{0.8952807659705883,0.6156984995294118,0.6081210191470588} .683 & \large\cellcolor[rgb]{0.61490285,0.649358983,0.8768415764999999} .540 & \large\cellcolor[rgb]{0.7048055483372548,0.7703792543686274,0.9677723612431373} .552 & \large\cellcolor[rgb]{0.6643022236588235,0.7198559149882353,0.9347936312705882} 4.56 \\
HUQ-SATRMD & \large\cellcolor[rgb]{0.852836579,0.50777808,0.575116406} \textbf{.395} & \large\cellcolor[rgb]{0.852836579,0.50777808,0.575116406} \textbf{.187} & \large\cellcolor[rgb]{0.852836579,0.50777808,0.575116406} \textbf{.486} & \large\cellcolor[rgb]{0.8415278407803921,0.8950213134450979,0.9948842140921569} .297 & \large\cellcolor[rgb]{0.9441952453705882,0.708851458745098,0.6639489555019608} .126 & \large\cellcolor[rgb]{0.9841016995,0.8604220499999999,0.8061464956666666} \underline{.048} & \large\cellcolor[rgb]{0.9028614815235294,0.6299065681294118,0.6152808287} .351 & \large\cellcolor[rgb]{0.8704786593764706,0.5611201626352941,0.5878720995529412} \underline{.211} & \large\cellcolor[rgb]{0.61490285,0.649358983,0.8768415764999999} .386 & \large\cellcolor[rgb]{0.8594926464901961,0.905996446872549,0.9888280806960784} .506 & \large\cellcolor[rgb]{0.9727701494549019,0.8993028702666667,0.8615527086490196} .308 & \large\cellcolor[rgb]{0.9218513348607843,0.6650340647294117,0.635032405109804} \underline{.450} & \large\cellcolor[rgb]{0.852836579,0.50777808,0.575116406} \textbf{.653} & \large\cellcolor[rgb]{0.9460687713941176,0.7126943685490196,0.6666446363803922} .646 & \large\cellcolor[rgb]{0.9377786937156862,0.9301210794313726,0.9257150330490196} .592 & \large\cellcolor[rgb]{0.8204138911862745,0.8803757532058823,0.9989228874411764} .609 & \large\cellcolor[rgb]{0.9646785691470589,0.756126767372549,0.7003363715784314} 2.94 \\\midrule
SATMD+MSP & \large\cellcolor[rgb]{0.9729271893941176,0.7797824782294118,0.7207566595882353} .234 & \large\cellcolor[rgb]{0.9844312501823529,0.8554192419058824,0.7999502522156863} .086 & \large\cellcolor[rgb]{0.7853082184392157,0.8520545809019608,0.9985974155294117} .377 & \large\cellcolor[rgb]{0.852836579,0.50777808,0.575116406} \textbf{.420} & \large\cellcolor[rgb]{0.9176723556764705,0.9302569986470588,0.9494852049705882} .094 & \large\cellcolor[rgb]{0.852836579,0.50777808,0.575116406} \textbf{.074} & \large\cellcolor[rgb]{0.852836579,0.50777808,0.575116406} \underline{.371} & \large\cellcolor[rgb]{0.9305268001470588,0.6814578817647059,0.6453140635882353} .203 & \large\cellcolor[rgb]{0.852836579,0.50777808,0.575116406} \textbf{.493} & \large\cellcolor[rgb]{0.9627817115,0.9127586490352941,0.8855681539058824} .527 & \large\cellcolor[rgb]{0.852836579,0.50777808,0.575116406} \textbf{.361} & \large\cellcolor[rgb]{0.852836579,0.50777808,0.575116406} \textbf{.466} & \large\cellcolor[rgb]{0.61490285,0.649358983,0.8768415764999999} .178 & \large\cellcolor[rgb]{0.852836579,0.50777808,0.575116406} \textbf{.708} & \large\cellcolor[rgb]{0.9756268974411765,0.7893996947941176,0.729703902882353} \underline{.618} & \large\cellcolor[rgb]{0.852836579,0.50777808,0.575116406} \textbf{.836} & \large\cellcolor[rgb]{0.852836579,0.50777808,0.575116406} \textbf{2.50} \\
SATRMD+MSP & \large\cellcolor[rgb]{0.8763519705509804,0.5787878133490196,0.5921289546450981} .372 & \large\cellcolor[rgb]{0.8704786593764706,0.5611201626352941,0.5878720995529412} .179 & \large\cellcolor[rgb]{0.8123516188058824,0.8741592436176471,0.9993597327901961} .383 & \large\cellcolor[rgb]{0.8979687697431373,0.6209226426705883,0.6104148745666667} \underline{.408} & \large\cellcolor[rgb]{0.8790560841823529,0.5840607685176471,0.5944135352882353} \underline{.135} & \large\cellcolor[rgb]{0.7798733627843137,0.8473838640392157,0.9980376545882352} .016 & \large\cellcolor[rgb]{0.852836579,0.50777808,0.575116406} \textbf{.372} & \large\cellcolor[rgb]{0.9367011412764705,0.6934798195294117,0.6531662319882353} .202 & \large\cellcolor[rgb]{0.9791396989627451,0.8021675484411765,0.7416485506941177} .466 & \large\cellcolor[rgb]{0.852836579,0.50777808,0.575116406} \textbf{.575} & \large\cellcolor[rgb]{0.8952807659705883,0.6156984995294118,0.6081210191470588} \underline{.353} & \large\cellcolor[rgb]{0.9844470791666666,0.8397397817137255,0.7814061455764706} .419 & \large\cellcolor[rgb]{0.9747269947588235,0.7861939559392157,0.7267214884509805} .542 & \large\cellcolor[rgb]{0.8616576191882352,0.5344491213176471,0.5814942527764706} \underline{.702} & \large\cellcolor[rgb]{0.852836579,0.50777808,0.575116406} \textbf{.642} & \large\cellcolor[rgb]{0.8952807659705883,0.6156984995294118,0.6081210191470588} \underline{.816} & \large\cellcolor[rgb]{0.8704786593764706,0.5611201626352941,0.5878720995529412} \underline{2.56} \\
\bottomrule
\end{tabular}
}\caption{\label{tab:llama_agg_results}Performance of various versions of the proposed supervised methods. PRR$\uparrow$ for Llama 8b v3.1 model for various tasks for the considered sequence-level aggregation methods. Warmer color indicates better results.}\end{table*}

%% file: llama.tex
\begin{table*}[t!] \resizebox{\textwidth}{!}{
\begin{tabular}{l||cc|cc|cc|cc|cc|c|c|c|c|c|c|c}
\toprule
\multirow{2}{*}{\textbf{UQ Method}} & \multicolumn{2}{c|}{\textbf{XSUM}} & \multicolumn{2}{c|}{\textbf{SamSum}} & \multicolumn{2}{c|}{\textbf{CNN}}& \multicolumn{2}{c|}{\textbf{PubMedQA}} & \multicolumn{2}{c|}{\textbf{MedQUAD}} & \multicolumn{1}{c|}{\textbf{TruthfulQA}}& \multicolumn{1}{c|}{\textbf{CoQA}} & \multicolumn{1}{c|}{\textbf{SciQ}} & \multicolumn{1}{c|}{\textbf{TriviaQA}} & \multicolumn{1}{c|}{\textbf{GSM8k}} & \multicolumn{1}{c|}{\textbf{MMLU}} & \multirow{2}{*}{\textbf{Mean Rank}} \\ \cline{2-17}
    & \textbf{ROUGE-L} & \textbf{AlignScore} & \textbf{ROUGE-L} & \textbf{AlignScore} & \textbf{ROUGE-L} & \textbf{AlignScore} & \textbf{ROUGE-L} & \textbf{AlignScore} & \textbf{ROUGE-L} & \textbf{AlignScore} & \textbf{AlignScore} & \textbf{AlignScore} & \textbf{AlignScore}& \textbf{AlignScore} & \textbf{Accuracy}& \textbf{Accuracy}& \\
\bottomrule
\end{tabular}
}\caption{\label{tab:llama_results} Main results on selective generation tasks. PRR$\uparrow$ for Llama 8b v3.1 model for various tasks for the considered sequence-level methods. Warmer color indicates better results.}\end{table*}

%% file: generalization_llama_v2.tex
\begin{table}[t] \centering
\resizebox{\linewidth}{!}{\begin{tabular}{l||cc|cc|c|c|c|c}
\toprule
\multirow{2}{*}{\textbf{UQ Method}} & \multicolumn{2}{c|}{\textbf{SamSum}} & \multicolumn{2}{c|}{\textbf{MedQUAD}} & \multicolumn{1}{c|}{\textbf{TruthfulQA}} & \multicolumn{1}{c|}{\textbf{SciQ}} & \multicolumn{1}{c|}{\textbf{MMLU}} & \multirow{2}{*}{\textbf{Mean Rank}} \\ \cline{2-8}
    & \textbf{ROUGE-L} & \textbf{AlignScore}& \textbf{ROUGE-L} & \textbf{AlignScore} & \textbf{AlignScore} & \textbf{AlignScore} & \textbf{Accuracy} \\
\midrule
Maximum Sequence Probability & \large\cellcolor[rgb]{0.852836579,0.50777808,0.575116406} \textbf{.452} & \large\cellcolor[rgb]{0.9834812190705882,0.867835001309804,0.8154382697392157} .218 & \large\cellcolor[rgb]{0.8734190061058824,0.5700105097411765,0.5899980484784314} \underline{.297} & \large\cellcolor[rgb]{0.852836579,0.50777808,0.575116406} \textbf{.356} & \large\cellcolor[rgb]{0.8675383126470588,0.5522298155294117,0.585746150627451} \underline{.277} & \large\cellcolor[rgb]{0.9261890675039215,0.6732459732470588,0.6401732343490196} .582 & \large\cellcolor[rgb]{0.9678868848333333,0.9061183506196079,0.873578236792157} .405 & \large\cellcolor[rgb]{0.9261890675039215,0.6732459732470588,0.6401732343490196} \underline{2.29} \\\midrule
DegMat NLI Score Entail. & \large\cellcolor[rgb]{0.9236824528058823,0.9312362411882353,0.942770235182353} .250 & \large\cellcolor[rgb]{0.9847608008647059,0.8504164338117647,0.7937540087647059} .226 & \large\cellcolor[rgb]{0.61490285,0.649358983,0.8768415764999999} .066 & \large\cellcolor[rgb]{0.9024823794117647,0.9258330802784314,0.9630825372156863} .162 & \large\cellcolor[rgb]{0.9514243351588236,0.9223978252941176,0.9059849168705882} .156 & \large\cellcolor[rgb]{0.9818859091411765,0.8745427552862746,0.8247100216450981} .446 & \large\cellcolor[rgb]{0.8569262456745098,0.9044285706686275,0.989693242609804} .224 & \large\cellcolor[rgb]{0.8620206859411765,0.9074551963235293,0.9878254853235294} 5.00 \\
SAR & \large\cellcolor[rgb]{0.9763803588352942,0.8914823988,0.8493228856529411} .300 & \large\cellcolor[rgb]{0.9829494490941176,0.8700709193019608,0.8185288537078431} .217 & \large\cellcolor[rgb]{0.9053078374137256,0.6343985308588236,0.6177138057666667} .286 & \large\cellcolor[rgb]{0.9479408841470589,0.9249530282941176,0.9117495381470588} .192 & \large\cellcolor[rgb]{0.8440942415960784,0.8965891896490196,0.9940190521784313} .105 & \large\cellcolor[rgb]{0.9802905992117648,0.8812505092627452,0.8339817735509804} .440 & \large\cellcolor[rgb]{0.9003004236470589,0.9251791607803921,0.9650037801960785} .284 & \large\cellcolor[rgb]{0.8956961428039216,0.9233751040392157,0.9683204633137255} 4.71 \\
Semantic Entropy & \large\cellcolor[rgb]{0.9813541391647058,0.8767786732784314,0.8278006056137255} .311 & \large\cellcolor[rgb]{0.9763803588352942,0.8914823988,0.8493228856529411} .206 & \large\cellcolor[rgb]{0.6355521478235294,0.6800053309882352,0.9035475637176471} .075 & \large\cellcolor[rgb]{0.61490285,0.649358983,0.8768415764999999} .007 & \large\cellcolor[rgb]{0.9717157648333333,0.9011381268078431,0.8645857989568628} .171 & \large\cellcolor[rgb]{0.9845960255235294,0.8529178378588236,0.7968521304901961} .466 & \large\cellcolor[rgb]{0.8543598448588235,0.9028606944647058,0.9905584045235294} .220 & \large\cellcolor[rgb]{0.8256989195784313,0.8840607433235294,0.9979455750647059} 5.29 \\\midrule
Factoscope & \large\cellcolor[rgb]{0.61490285,0.649358983,0.8768415764999999} .061 & \large\cellcolor[rgb]{0.61490285,0.649358983,0.8768415764999999} .037 & \large\cellcolor[rgb]{0.6594161915960784,0.7133025255607843,0.9299287241019608} .084 & \large\cellcolor[rgb]{0.7880256462666666,0.8543899393333334,0.9988772960000001} .101 & \large\cellcolor[rgb]{0.6944259359764706,0.7581492177882353,0.9606867415411764} .045 & \large\cellcolor[rgb]{0.9727701494549019,0.8993028702666667,0.8615527086490196} .420 & \large\cellcolor[rgb]{0.7365350864941176,0.8055387188078431,0.9853167941313725} .071 & \large\cellcolor[rgb]{0.61490285,0.649358983,0.8768415764999999} 7.00 \\
SAPLMA & \large\cellcolor[rgb]{0.9112102024705883,0.9284487582705883,0.9553975652941177} .241 & \large\cellcolor[rgb]{0.6996157421568627,0.7642642360784314,0.9642295513921568} .075 & \large\cellcolor[rgb]{0.6449978024117646,0.6934178380941176,0.9144631795921568} .079 & \large\cellcolor[rgb]{0.8440942415960784,0.8965891896490196,0.9940190521784313} .129 & \large\cellcolor[rgb]{0.61490285,0.649358983,0.8768415764999999} .009 & \large\cellcolor[rgb]{0.61490285,0.649358983,0.8768415764999999} .091 & \large\cellcolor[rgb]{0.61490285,0.649358983,0.8768415764999999} -.097 & \large\cellcolor[rgb]{0.61490285,0.649358983,0.8768415764999999} 7.00 \\\midrule
HUQ-SATRMD & \large\cellcolor[rgb]{0.9196824685392158,0.6609281104705882,0.632461990490196} \underline{.413} & \large\cellcolor[rgb]{0.9846442845,0.8424908735411765,0.7844876631294118} \underline{.229} & \large\cellcolor[rgb]{0.8844643114450981,0.5946066788549021,0.5989826965745099} .293 & \large\cellcolor[rgb]{0.852836579,0.50777808,0.575116406} \underline{.355} & \large\cellcolor[rgb]{0.852836579,0.50777808,0.575116406} \textbf{.283} & \large\cellcolor[rgb]{0.852836579,0.50777808,0.575116406} \textbf{.644} & \large\cellcolor[rgb]{0.852836579,0.50777808,0.575116406} \textbf{.770} & \large\cellcolor[rgb]{0.852836579,0.50777808,0.575116406} \textbf{1.71} \\
SATRMD+MSP & \large\cellcolor[rgb]{0.8594926464901961,0.905996446872549,0.9888280806960784} .207 & \large\cellcolor[rgb]{0.852836579,0.50777808,0.575116406} \textbf{.314} & \large\cellcolor[rgb]{0.852836579,0.50777808,0.575116406} \textbf{.304} & \large\cellcolor[rgb]{0.9460687713941176,0.7126943685490196,0.6666446363803922} .304 & \large\cellcolor[rgb]{0.8840171821764706,0.9185176097647059,0.976244109117647} .122 & \large\cellcolor[rgb]{0.9102005491941176,0.643382456317647,0.6225797599} \underline{.598} & \large\cellcolor[rgb]{0.9218513348607843,0.6650340647294117,0.635032405109804} \underline{.681} & \large\cellcolor[rgb]{0.9774267028058823,0.7958111725039216,0.7356687317450981} 3.00 \\
\bottomrule
\end{tabular}
}\caption{\label{tab:llama_gen_results}  Out-of-domain generalization. PRR$\uparrow$ for Llama 8b v3.1 for selective generation tasks for the considered sequence-level methods in the out-of-domain setting. Warmer color indicates better results.}\end{table}

%% file: factchecking.tex
\begin{table}[t]\footnotesize
\centering\resizebox{0.3\textwidth}{!}{\begin{tabular}{l||cc}
\toprule
\multirow{2}{*}{\textbf{UQ Method}} & \multicolumn{2}{c}{\textbf{Mistral 7b}} \\ \cline{2-3}
    & \textbf{ROC-AUC} & \textbf{PR-AUC} \\
\midrule
Maximum Claim Probability & \small\cellcolor[rgb]{0.9337138175431372,0.9321882998862745,0.9313012310098039} .620 & \small\cellcolor[rgb]{0.8910245585529412,0.9214321063294117,0.9714899216352941} .271 \\
P(True) & \small\cellcolor[rgb]{0.9653342981666666,0.909438499827451,0.8795731953490196} .638 & \small\cellcolor[rgb]{0.9024823794117647,0.9258330802784314,0.9630825372156863} .276 \\
CCP & \small\cellcolor[rgb]{0.9367011412764705,0.6934798195294117,0.6531662319882353} .716 & \small\cellcolor[rgb]{0.9218513348607843,0.6650340647294117,0.635032405109804} .388 \\
SAPLMA & \small\cellcolor[rgb]{0.61490285,0.649358983,0.8768415764999999} .489 & \small\cellcolor[rgb]{0.61490285,0.649358983,0.8768415764999999} .166 \\ \midrule
SATRMD & \small\cellcolor[rgb]{0.974575254145098,0.8953926345333334,0.8554377971509803} .647 & \small\cellcolor[rgb]{0.9003004236470589,0.9251791607803921,0.9650037801960785} .275 \\
HUQ-SATRMD & \small\cellcolor[rgb]{0.852836579,0.50777808,0.575116406} \textbf{.750} & \small\cellcolor[rgb]{0.864597965917647,0.5433394684235294,0.5836202017019608} \underline{.410} \\
SATRMD+CCP & \small\cellcolor[rgb]{0.8817601978137255,0.5893337236862746,0.5966981159313726} \underline{.739} & \small\cellcolor[rgb]{0.852836579,0.50777808,0.575116406} \textbf{.414} \\
\bottomrule
\end{tabular}
}\caption{\label{tab:factcheking_results} Results in fact-checking. ROC-AUC$\uparrow$ and PR-AUC$\uparrow$ for the Mistral 7b v0.1 Instruct model for fact-checking for the considered claim-level methods. Warmer color indicates better results.}\end{table}

%% file: gemma_aggregations.tex
\begin{table*}[!ht] \resizebox{\textwidth}{!}{
\begin{tabular}{l||cc|cc|cc|cc|cc|c|c|c|c|c|c|c}
\toprule
\multirow{2}{*}{\textbf{UQ Method}} & \multicolumn{2}{c|}{\textbf{XSUM}} & \multicolumn{2}{c|}{\textbf{SamSum}} & \multicolumn{2}{c|}{\textbf{CNN}}& \multicolumn{2}{c|}{\textbf{PubMedQA}} & \multicolumn{2}{c|}{\textbf{MedQUAD}} & \multicolumn{1}{c|}{\textbf{TruthfulQA}}& \multicolumn{1}{c|}{\textbf{CoQA}} & \multicolumn{1}{c|}{\textbf{SciQ}} & \multicolumn{1}{c|}{\textbf{TriviaQA}} & \multicolumn{1}{c|}{\textbf{GSM8k}} & \multicolumn{1}{c|}{\textbf{MMLU}} & \multirow{2}{*}{\textbf{Mean Rank}} \\ \cline{2-17}
    & \textbf{ROUGE-L} & \textbf{AlignScore} & \textbf{ROUGE-L} & \textbf{AlignScore} & \textbf{ROUGE-L} & \textbf{AlignScore} & \textbf{ROUGE-L} & \textbf{AlignScore} & \textbf{ROUGE-L} & \textbf{AlignScore} & \textbf{AlignScore} & \textbf{AlignScore} & \textbf{AlignScore}& \textbf{AlignScore} & \textbf{Accuracy}& \textbf{Accuracy}& \\
\bottomrule
\end{tabular}
}\caption{\label{tab:gemma_agg_results} Performance of various versions of the proposed supervised methods. PRR$\uparrow$ for Gemma 9b v2 model for various tasks for the considered sequence-level aggregation methods. Warmer color indicates better results.}\end{table*}

%% file: gemma.tex
\begin{table*}[!ht] \resizebox{\textwidth}{!}{\begin{tabular}{l||cc|cc|cc|cc|cc|c|c|c|c|c|c|c}
\toprule
\multirow{2}{*}{\textbf{UQ Method}} & \multicolumn{2}{c|}{\textbf{XSUM}} & \multicolumn{2}{c|}{\textbf{SamSum}} & \multicolumn{2}{c|}{\textbf{CNN}}& \multicolumn{2}{c|}{\textbf{PubMedQA}} & \multicolumn{2}{c|}{\textbf{MedQUAD}} & \multicolumn{1}{c|}{\textbf{TruthfulQA}}& \multicolumn{1}{c|}{\textbf{CoQA}} & \multicolumn{1}{c|}{\textbf{SciQ}} & \multicolumn{1}{c|}{\textbf{TriviaQA}} & \multicolumn{1}{c|}{\textbf{GSM8k}} & \multicolumn{1}{c|}{\textbf{MMLU}} & \multirow{2}{*}{\textbf{Mean Rank}} \\ \cline{2-17}
    & \textbf{ROUGE-L} & \textbf{AlignScore} & \textbf{ROUGE-L} & \textbf{AlignScore} & \textbf{ROUGE-L} & \textbf{AlignScore} & \textbf{ROUGE-L} & \textbf{AlignScore} & \textbf{ROUGE-L} & \textbf{AlignScore} & \textbf{AlignScore} & \textbf{AlignScore} & \textbf{AlignScore}& \textbf{AlignScore} & \textbf{Accuracy}& \textbf{Accuracy}& \\
\bottomrule
\end{tabular}
}\caption{\label{tab:gemma_results} Main results on selective generation tasks. PRR$\uparrow$ for Gemma 9b v2 model for various tasks for the considered sequence-level methods. Warmer color indicates better results.}\end{table*}

%% file: comp_time.tex
\begin{table}[h]\scriptsize  \centering\begin{tabular}{l|c|c}
\toprule
\textbf{UQ Method} & \textbf{\multirowcell{Runtime \\ per batch}} & \textbf{\multirowcell{Overhead}} \\
\midrule

MSP & $2.10$\tiny{$\pm$$1.31$} & - \\\midrule
DegMat NLI Score Entail. & $9.47$\tiny{$\pm$$3.41$} & 350\% \\
Lexical Similarity ROUGE-L & $8.69$\tiny{$\pm$$3.31$} & 315\% \\
Semantic Entropy & $9.47$\tiny{$\pm$$3.41$} & 350\% \\
SAR & $16.89$\tiny{$\pm$$6.85$} & 700\% \\
\midrule
SAPLMA & $2.10$\tiny{$\pm$$1.31$} & \textbf{0.04}\% \\
Factoscope & $8.45$\tiny{$\pm$$5.92$} & 300\% \\
\midrule
HUQ-SATRMD & $2.21$\tiny{$\pm$$1.36$} & \underline{5.30}\% \\
SATRMD+MSP & $2.26$\tiny{$\pm$$1.38$} & 7.61 \%\\

\bottomrule
\end{tabular}
\caption{\label{tab:comp_efficiency} The evaluation of the inference runtime of UQ methods measured on all test instances from all datasets with predictions from Llama 8b v3.1. The best results are in bold. The second best results are underlined.}
\end{table}

%% file: datasets.tex
\begin{table}[h] \footnotesize  \centering\resizebox{0.48\textwidth}{!}{\begin{tabular}{c|l|c|c|c}
\toprule
\textbf{Task} & \textbf{Dataset} & \textbf{N-shot} & \multirowcell{\textbf{Train texts} \\ \textbf{for STMD}} & \multirowcell{\textbf{Evaluation} \\ \textbf{texts}} \\
\midrule
\multirow{3}{*}{\multirowcell{Text \\ Summarization}} & CNN/DailyMail & 0 & 2,000 & 2,000 \\
& XSum & 0 & 2,000 & 2,000 \\
& SamSum & 0 & 2,000 & 819 \\
\midrule
\multirow{4}{*}{\multirowcell{QA \\ Long answer}} & PubMedQA & 0 & 2,000 & 2,000 \\
& MedQUAD & 5 & 2,000 & 2,000 \\
& TruthfulQA & 5 & 408 & 409 \\
& GSM8k & 5 & 2,000 & 1,319 \\
\midrule
\multirow{4}{*}{\multirowcell{QA \\ Short answer}} & SciQ & 0 & 5,000 & 1,000 \\
& CoQA & \multirowcell{all preceding \\ questions} & 5,000 & 2,000 \\
& TriviaQA & 5 & 5,000 & 2,000 \\
\midrule
\multirow{1}{*}{\multirowcell{MCQA}} & MMLU & 5 & 5,000 & 2,000 \\
\bottomrule
\end{tabular}
}\caption{\label{tab:dataset_stat} The statistics of the datasets used for evaluation.}
\end{table}

%% file: llms_hp.tex
\begin{table*}[h] 
\centering\resizebox{\textwidth}{!}{
\begin{tabular}{l|c|c|c|c|c|c|c|c}
\toprule
\textbf{Dataset} & \textbf{Task} & \textbf{Max Input Length} & \textbf{Generation Length} & \textbf{Temperature} & \textbf{Top-p} & \textbf{Do Sample} & \textbf{Beams} & \textbf{Repetition Penalty} \\
\midrule
 XSum & \multirow{3}{*}{TS} & \multirow{11}{*}{-} & 56 & \multirow{11}{*}{1.0} & \multirow{11}{*}{1.0} & \multirow{11}{*}{False} & \multirow{11}{*}{1} & \multirow{11}{*}{1}  \\
 SamSum &  &  & 128 &  &  &  &  & \\
 CNN &  &  & 128 &  &  &  &  & \\

 PubMedQA & \multirow{4}{*}{\multirowcell{QA \\ Long answer}} & & 128 &  &  &  &  & \\
 MedQUAD & &  & 128 &  &  &  &  & \\
 TruthfulQA & &  & 128 &  &  &  &  & \\
 GSM8k &  &  & 256 &  &  &  &  & \\
 
 CoQA & \multirow{3}{*}{\multirowcell{QA \\ Short answer}} & & 20 & & & & & \\
 SciQ &  &  & 20 &  &  &  &  &  \\
 TriviQA &  &  & 20 &  &  &  &  &  \\
 MMLU &  \multirow{1}{*}{\multirowcell{MCQA}} &  & 3 &  &  &  &  &  \\

\bottomrule
\end{tabular}}
\caption{\label{tab:llm_hyperparameters} Text generation hyperparameters for all LLMs used in the experiments.}
\end{table*}

%% file: acl_latex.bbl
\begin{thebibliography}{65}
\providecommand{\natexlab}[1]{#1}

\bibitem[{Abacha and Demner{-}Fushman(2019)}]{medquad}
Asma~Ben Abacha and Dina Demner{-}Fushman. 2019.
\newblock \href {https://doi.org/10.1186/S12859-019-3119-4} {A question-entailment approach to question answering}.
\newblock \emph{{BMC} Bioinform.}, 20(1):511:1--511:23.

\bibitem[{Azaria and Mitchell(2023)}]{azaria-mitchell-2023-internal}
Amos Azaria and Tom Mitchell. 2023.
\newblock \href {https://doi.org/10.18653/v1/2023.findings-emnlp.68} {The internal state of an {LLM} knows when it{'}s lying}.
\newblock In \emph{Findings of the Association for Computational Linguistics: EMNLP 2023}, pages 967--976, Singapore. Association for Computational Linguistics.

\bibitem[{Baan et~al.(2023)Baan, Daheim, Ilia, Ulmer, Li, Fern{\'a}ndez, Plank, Sennrich, Zerva, and Aziz}]{baan2023uncertainty}
Joris Baan, Nico Daheim, Evgenia Ilia, Dennis Ulmer, Haau-Sing Li, Raquel Fern{\'a}ndez, Barbara Plank, Rico Sennrich, Chrysoula Zerva, and Wilker Aziz. 2023.
\newblock \href {https://arxiv.org/abs/2307.15703} {Uncertainty in natural language generation: From theory to applications}.
\newblock \emph{arXiv preprint arXiv:2307.15703}.

\bibitem[{CH-Wang et~al.(2024)CH-Wang, Van~Durme, Eisner, and Kedzie}]{ch-wang-etal-2024-androids}
Sky CH-Wang, Benjamin Van~Durme, Jason Eisner, and Chris Kedzie. 2024.
\newblock \href {https://aclanthology.org/2024.findings-acl.260} {Do androids know they{'}re only dreaming of electric sheep?}
\newblock In \emph{Findings of the Association for Computational Linguistics ACL 2024}, pages 4401--4420, Bangkok, Thailand and virtual meeting. Association for Computational Linguistics.

\bibitem[{Chan et~al.(2022)Chan, Leow, Bea, Cheng, Phoong, Hong, and Chen}]{mitigating_multicollinearity}
Jireh Chan, Steven Leow, Khean Bea, Wai~Khuen Cheng, Seuk~Wai Phoong, Zeng-Wei Hong, and Yen-Lin Chen. 2022.
\newblock \href {https://doi.org/10.3390/math10081283} {Mitigating the multicollinearity problem and its machine learning approach: A review}.
\newblock \emph{Mathematics}, 10:1283.

\bibitem[{Chen et~al.(2024)Chen, Liu, Chen, Gu, Wu, Tao, Fu, and Ye}]{chen2024eigenscore}
Chao Chen, Kai Liu, Ze~Chen, Yi~Gu, Yue Wu, Mingyuan Tao, Zhihang Fu, and Jieping Ye. 2024.
\newblock \href {https://openreview.net/forum?id=Zj12nzlQbz} {{INSIDE:} llms' internal states retain the power of hallucination detection}.
\newblock In \emph{The Twelfth International Conference on Learning Representations, {ICLR} 2024, Vienna, Austria, May 7-11, 2024}. OpenReview.net.

\bibitem[{Chen et~al.(2023)Chen, Fu, Yuan, Wen, Fan, Liu, Zhang, Li, and Xiao}]{chen2023hallucination}
Yuyan Chen, Qiang Fu, Yichen Yuan, Zhihao Wen, Ge~Fan, Dayiheng Liu, Dongmei Zhang, Zhixu Li, and Yanghua Xiao. 2023.
\newblock \href {https://arxiv.org/abs/2407.04121} {Hallucination detection: Robustly discerning reliable answers in large language models}.
\newblock In \emph{Proceedings of the 32nd ACM International Conference on Information and Knowledge Management}, pages 245--255.

\bibitem[{Cheng and Vlachos(2024)}]{cheng-vlachos-2024-measuring}
Julius Cheng and Andreas Vlachos. 2024.
\newblock \href {https://aclanthology.org/2024.eacl-long.129} {Measuring uncertainty in neural machine translation with similarity-sensitive entropy}.
\newblock In \emph{Proceedings of the 18th Conference of the European Chapter of the Association for Computational Linguistics (Volume 1: Long Papers)}, pages 2115--2128, St. Julian{'}s, Malta. Association for Computational Linguistics.

\bibitem[{Cobbe et~al.(2021)Cobbe, Kosaraju, Bavarian, Chen, Jun, Kaiser, Plappert, Tworek, Hilton, Nakano, Hesse, and Schulman}]{gsm8k}
Karl Cobbe, Vineet Kosaraju, Mohammad Bavarian, Mark Chen, Heewoo Jun, Lukasz Kaiser, Matthias Plappert, Jerry Tworek, Jacob Hilton, Reiichiro Nakano, Christopher Hesse, and John Schulman. 2021.
\newblock \href {https://arxiv.org/abs/2110.14168} {Training verifiers to solve math word problems}.
\newblock \emph{arXiv preprint arXiv:2110.14168}.

\bibitem[{Darrin et~al.(2023)Darrin, Piantanida, and Colombo}]{darrin-etal-2023-rainproof}
Maxime Darrin, Pablo Piantanida, and Pierre Colombo. 2023.
\newblock \href {https://doi.org/10.18653/v1/2023.emnlp-main.357} {{R}ain{P}roof: An umbrella to shield text generator from out-of-distribution data}.
\newblock In \emph{Proceedings of the 2023 Conference on Empirical Methods in Natural Language Processing}, pages 5831--5857, Singapore. Association for Computational Linguistics.

\bibitem[{Duan et~al.(2024)Duan, Cheng, Wang, Zavalny, Wang, Xu, Kailkhura, and Xu}]{duan-etal-2024-shifting}
Jinhao Duan, Hao Cheng, Shiqi Wang, Alex Zavalny, Chenan Wang, Renjing Xu, Bhavya Kailkhura, and Kaidi Xu. 2024.
\newblock \href {https://doi.org/10.18653/v1/2024.acl-long.276} {Shifting attention to relevance: Towards the predictive uncertainty quantification of free-form large language models}.
\newblock In \emph{Proceedings of the 62nd Annual Meeting of the Association for Computational Linguistics (Volume 1: Long Papers)}, pages 5050--5063, Bangkok, Thailand. Association for Computational Linguistics.

\bibitem[{Dubey et~al.(2024)Dubey, Jauhri, Pandey, Kadian, Al{-}Dahle, Letman, Mathur, Schelten, Yang, Fan, Goyal, Hartshorn, Yang, Mitra, Sravankumar, Korenev, Hinsvark, Rao, Zhang, Rodriguez, Gregerson, Spataru, Rozi{\`{e}}re, Biron, Tang, Chern, Caucheteux, Nayak, Bi, Marra, McConnell, Keller, Touret, Wu, Wong, Ferrer, Nikolaidis, Allonsius, Song, Pintz, Livshits, Esiobu, Choudhary, Mahajan, Garcia{-}Olano, Perino, Hupkes, Lakomkin, AlBadawy, Lobanova, Dinan, Smith, Radenovic, Zhang, Synnaeve, Lee, Anderson, Nail, Mialon, Pang, Cucurell, Nguyen, Korevaar, Xu, Touvron, Zarov, Ibarra, Kloumann, Misra, Evtimov, Copet, Lee, Geffert, Vranes, Park, Mahadeokar, Shah, van~der Linde, Billock, Hong, Lee, Fu, Chi, Huang, Liu, Wang, Yu, Bitton, Spisak, Park, Rocca, Johnstun, Saxe, Jia, Alwala, Upasani, Plawiak, Li, Heafield, Stone, and et~al.}]{llama3}
Abhimanyu Dubey, Abhinav Jauhri, Abhinav Pandey, Abhishek Kadian, Ahmad Al{-}Dahle, Aiesha Letman, Akhil Mathur, Alan Schelten, Amy Yang, Angela Fan, Anirudh Goyal, Anthony Hartshorn, Aobo Yang, Archi Mitra, Archie Sravankumar, Artem Korenev, Arthur Hinsvark, Arun Rao, Aston Zhang, Aur{\'{e}}lien Rodriguez, Austen Gregerson, Ava Spataru, Baptiste Rozi{\`{e}}re, Bethany Biron, Binh Tang, Bobbie Chern, Charlotte Caucheteux, Chaya Nayak, Chloe Bi, Chris Marra, Chris McConnell, Christian Keller, Christophe Touret, Chunyang Wu, Corinne Wong, Cristian~Canton Ferrer, Cyrus Nikolaidis, Damien Allonsius, Daniel Song, Danielle Pintz, Danny Livshits, David Esiobu, Dhruv Choudhary, Dhruv Mahajan, Diego Garcia{-}Olano, Diego Perino, Dieuwke Hupkes, Egor Lakomkin, Ehab AlBadawy, Elina Lobanova, Emily Dinan, Eric~Michael Smith, Filip Radenovic, Frank Zhang, Gabriel Synnaeve, Gabrielle Lee, Georgia~Lewis Anderson, Graeme Nail, Gr{\'{e}}goire Mialon, Guan Pang, Guillem Cucurell, Hailey Nguyen, Hannah Korevaar, Hu~Xu, Hugo
  Touvron, Iliyan Zarov, Imanol~Arrieta Ibarra, Isabel~M. Kloumann, Ishan Misra, Ivan Evtimov, Jade Copet, Jaewon Lee, Jan Geffert, Jana Vranes, Jason Park, Jay Mahadeokar, Jeet Shah, Jelmer van~der Linde, Jennifer Billock, Jenny Hong, Jenya Lee, Jeremy Fu, Jianfeng Chi, Jianyu Huang, Jiawen Liu, Jie Wang, Jiecao Yu, Joanna Bitton, Joe Spisak, Jongsoo Park, Joseph Rocca, Joshua Johnstun, Joshua Saxe, Junteng Jia, Kalyan~Vasuden Alwala, Kartikeya Upasani, Kate Plawiak, Ke~Li, Kenneth Heafield, Kevin Stone, and et~al. 2024.
\newblock \href {https://doi.org/10.48550/arXiv.2407.21783} {The llama 3 herd of models}.
\newblock \emph{arXiv preprint arXiv:2407.21783}.

\bibitem[{Dziri et~al.(2022)Dziri, Milton, Yu, Zaiane, and Reddy}]{dziri-etal-2022-origin}
Nouha Dziri, Sivan Milton, Mo~Yu, Osmar Zaiane, and Siva Reddy. 2022.
\newblock \href {https://doi.org/10.18653/v1/2022.naacl-main.387} {On the origin of hallucinations in conversational models: Is it the datasets or the models?}
\newblock In \emph{Proceedings of the 2022 Conference of the North American Chapter of the Association for Computational Linguistics: Human Language Technologies}, pages 5271--5285, Seattle, United States. Association for Computational Linguistics.

\bibitem[{Fadeeva et~al.(2024)Fadeeva, Rubashevskii, Shelmanov, Petrakov, Li, Mubarak, Tsymbalov, Kuzmin, Panchenko, Baldwin, Nakov, and Panov}]{fadeeva2024factchecking}
Ekaterina Fadeeva, Aleksandr Rubashevskii, Artem Shelmanov, Sergey Petrakov, Haonan Li, Hamdy Mubarak, Evgenii Tsymbalov, Gleb Kuzmin, Alexander Panchenko, Timothy Baldwin, Preslav Nakov, and Maxim Panov. 2024.
\newblock \href {https://doi.org/10.48550/arXiv.2403.04696} {Fact-checking the output of large language models via token-level uncertainty quantification}.
\newblock In \emph{Findings of the Association for Computational Linguistics: ACL 2024}. Association for Computational Linguistics.

\bibitem[{Fadeeva et~al.(2023)Fadeeva, Vashurin, Tsvigun, Vazhentsev, Petrakov, Fedyanin, Vasilev, Goncharova, Panchenko, Panov, Baldwin, and Shelmanov}]{fadeeva2023lm}
Ekaterina Fadeeva, Roman Vashurin, Akim Tsvigun, Artem Vazhentsev, Sergey Petrakov, Kirill Fedyanin, Daniil Vasilev, Elizaveta Goncharova, Alexander Panchenko, Maxim Panov, Timothy Baldwin, and Artem Shelmanov. 2023.
\newblock \href {https://doi.org/10.48550/arXiv.2311.07383} {{LM-Polygraph}: Uncertainty estimation for language models}.
\newblock In \emph{Proceedings of the 2023 Conference on Empirical Methods in Natural Language Processing: System Demonstrations}, pages 446--461.

\bibitem[{Farquhar et~al.(2024)Farquhar, Kossen, Kuhn, and Gal}]{farquhar2024detecting}
Sebastian Farquhar, Jannik Kossen, Lorenz Kuhn, and Yarin Gal. 2024.
\newblock \href {https://www.nature.com/articles/s41586-024-07421-0} {Detecting hallucinations in large language models using semantic entropy}.
\newblock \emph{Nature}, 630(8017):625--630.

\bibitem[{Feng et~al.(2024)Feng, Shi, Wang, Ding, Balachandran, and Tsvetkov}]{feng-etal-2024-dont}
Shangbin Feng, Weijia Shi, Yike Wang, Wenxuan Ding, Vidhisha Balachandran, and Yulia Tsvetkov. 2024.
\newblock \href {https://doi.org/10.18653/v1/2024.acl-long.786} {Don{'}t hallucinate, abstain: Identifying {LLM} knowledge gaps via multi-{LLM} collaboration}.
\newblock In \emph{Proceedings of the 62nd Annual Meeting of the Association for Computational Linguistics (Volume 1: Long Papers)}, pages 14664--14690, Bangkok, Thailand. Association for Computational Linguistics.

\bibitem[{Fomicheva et~al.(2020)Fomicheva, Sun, Yankovskaya, Blain, Guzm{\'a}n, Fishel, Aletras, Chaudhary, and Specia}]{fomicheva-etal-2020-unsupervised}
Marina Fomicheva, Shuo Sun, Lisa Yankovskaya, Fr{\'e}d{\'e}ric Blain, Francisco Guzm{\'a}n, Mark Fishel, Nikolaos Aletras, Vishrav Chaudhary, and Lucia Specia. 2020.
\newblock \href {https://doi.org/10.1162/tacl_a_00330} {Unsupervised quality estimation for neural machine translation}.
\newblock \emph{Transactions of the Association for Computational Linguistics}, 8:539--555.

\bibitem[{Gal and Ghahramani(2016)}]{gal2016dropout}
Yarin Gal and Zoubin Ghahramani. 2016.
\newblock \href {https://proceedings.mlr.press/v48/gal16.html} {Dropout as a {Bayesian} approximation: Representing model uncertainty in deep learning}.
\newblock In \emph{Proceedings of The 33rd International Conference on Machine Learning}, volume~48 of \emph{Proceedings of Machine Learning Research}, pages 1050--1059, New York, New York, USA. PMLR.

\bibitem[{Geng et~al.(2024)Geng, Cai, Wang, Koeppl, Nakov, and Gurevych}]{geng2023survey}
Jiahui Geng, Fengyu Cai, Yuxia Wang, Heinz Koeppl, Preslav Nakov, and Iryna Gurevych. 2024.
\newblock \href {https://doi.org/10.18653/v1/2024.naacl-long.366} {A survey of confidence estimation and calibration in large language models}.
\newblock In \emph{Proceedings of the 2024 Conference of the North American Chapter of the Association for Computational Linguistics: Human Language Technologies (Volume 1: Long Papers)}, pages 6577--6595, Mexico City, Mexico. Association for Computational Linguistics.

\bibitem[{Gliwa et~al.(2019)Gliwa, Mochol, Biesek, and Wawer}]{gliwa-etal-2019-samsum}
Bogdan Gliwa, Iwona Mochol, Maciej Biesek, and Aleksander Wawer. 2019.
\newblock \href {https://doi.org/10.18653/v1/D19-5409} {{SAMS}um corpus: A human-annotated dialogue dataset for abstractive summarization}.
\newblock In \emph{Proceedings of the 2nd Workshop on New Frontiers in Summarization}, pages 70--79, Hong Kong, China. Association for Computational Linguistics.

\bibitem[{He et~al.(2024{\natexlab{a}})He, Yu, Lei, Lu, and Chen}]{he-etal-2024-uncertainty}
Jianfeng He, Linlin Yu, Shuo Lei, Chang-Tien Lu, and Feng Chen. 2024{\natexlab{a}}.
\newblock \href {https://doi.org/10.18653/v1/2024.findings-naacl.180} {Uncertainty estimation on sequential labeling via uncertainty transmission}.
\newblock In \emph{Findings of the Association for Computational Linguistics: NAACL 2024}, pages 2823--2835, Mexico City, Mexico. Association for Computational Linguistics.

\bibitem[{He et~al.(2020)He, Zhang, Lei, Chen, Chen, Alhamadani, Xiao, and Lu}]{he2020towards}
Jianfeng He, Xuchao Zhang, Shuo Lei, Zhiqian Chen, Fanglan Chen, Abdulaziz Alhamadani, Bei Xiao, and Chang{-}Tien Lu. 2020.
\newblock \href {https://doi.org/10.18653/v1/2020.emnlp-main.671} {Towards more accurate uncertainty estimation in text classification}.
\newblock In \emph{Proceedings of the 2020 Conference on Empirical Methods in Natural Language Processing, {EMNLP} 2020, Online, November 16-20, 2020}, pages 8362--8372. Association for Computational Linguistics.

\bibitem[{He et~al.(2024{\natexlab{b}})He, Gong, Lin, Wei, Zhao, and Chen}]{he-etal-2024-llm}
Jinwen He, Yujia Gong, Zijin Lin, Cheng{'}an Wei, Yue Zhao, and Kai Chen. 2024{\natexlab{b}}.
\newblock \href {https://doi.org/10.18653/v1/2024.findings-acl.608} {{LLM} factoscope: Uncovering {LLM}s{'} factual discernment through measuring inner states}.
\newblock In \emph{Findings of the Association for Computational Linguistics ACL 2024}, pages 10218--10230, Bangkok, Thailand and virtual meeting. Association for Computational Linguistics.

\bibitem[{Hendrycks et~al.(2021)Hendrycks, Burns, Basart, Zou, Mazeika, Song, and Steinhardt}]{mmlu}
Dan Hendrycks, Collin Burns, Steven Basart, Andy Zou, Mantas Mazeika, Dawn Song, and Jacob Steinhardt. 2021.
\newblock \href {https://openreview.net/forum?id=d7KBjmI3GmQ} {Measuring massive multitask language understanding}.
\newblock In \emph{9th International Conference on Learning Representations, {ICLR} 2021, Virtual Event, Austria, May 3-7, 2021}. OpenReview.net.

\bibitem[{Jiang et~al.(2023)Jiang, Sablayrolles, Mensch, Bamford, Chaplot, de~Las~Casas, Bressand, Lengyel, Lample, Saulnier, Lavaud, Lachaux, Stock, Scao, Lavril, Wang, Lacroix, and Sayed}]{mistral}
Albert~Q. Jiang, Alexandre Sablayrolles, Arthur Mensch, Chris Bamford, Devendra~Singh Chaplot, Diego de~Las~Casas, Florian Bressand, Gianna Lengyel, Guillaume Lample, Lucile Saulnier, L{\'{e}}lio~Renard Lavaud, Marie{-}Anne Lachaux, Pierre Stock, Teven~Le Scao, Thibaut Lavril, Thomas Wang, Timoth{\'{e}}e Lacroix, and William~El Sayed. 2023.
\newblock \href {https://doi.org/10.48550/arXiv.2310.06825} {Mistral 7b}.
\newblock \emph{arXiv preprint arXiv:2310.06825}.

\bibitem[{Jin et~al.(2019)Jin, Dhingra, Liu, Cohen, and Lu}]{jin-etal-2019-pubmedqa}
Qiao Jin, Bhuwan Dhingra, Zhengping Liu, William Cohen, and Xinghua Lu. 2019.
\newblock \href {https://doi.org/10.18653/v1/D19-1259} {{P}ub{M}ed{QA}: A dataset for biomedical research question answering}.
\newblock In \emph{Proceedings of the 2019 Conference on Empirical Methods in Natural Language Processing and the 9th International Joint Conference on Natural Language Processing (EMNLP-IJCNLP)}, pages 2567--2577, Hong Kong, China. Association for Computational Linguistics.

\bibitem[{Joshi et~al.(2017)Joshi, Choi, Weld, and Zettlemoyer}]{joshi-etal-2017-triviaqa}
Mandar Joshi, Eunsol Choi, Daniel Weld, and Luke Zettlemoyer. 2017.
\newblock \href {https://doi.org/10.18653/v1/P17-1147} {{T}rivia{QA}: A large scale distantly supervised challenge dataset for reading comprehension}.
\newblock In \emph{Proceedings of the 55th Annual Meeting of the Association for Computational Linguistics (Volume 1: Long Papers)}, pages 1601--1611, Vancouver, Canada. Association for Computational Linguistics.

\bibitem[{Kotelevskii et~al.(2022)Kotelevskii, Artemenkov, Fedyanin, Noskov, Fishkov, Shelmanov, Vazhentsev, Petiushko, and Panov}]{nuq_kotelevskii}
Nikita Kotelevskii, Aleksandr Artemenkov, Kirill Fedyanin, Fedor Noskov, Alexander Fishkov, Artem Shelmanov, Artem Vazhentsev, Aleksandr Petiushko, and Maxim Panov. 2022.
\newblock \href {https://openreview.net/forum?id=v6NNlubbSQ} {Nonparametric uncertainty quantification for single deterministic neural network}.
\newblock In \emph{Advances in Neural Information Processing Systems}.

\bibitem[{Kuhn et~al.(2023)Kuhn, Gal, and Farquhar}]{kuhn2023semantic}
Lorenz Kuhn, Yarin Gal, and Sebastian Farquhar. 2023.
\newblock \href {https://openreview.net/pdf?id=VD-AYtP0dve} {Semantic uncertainty: Linguistic invariances for uncertainty estimation in natural language generation}.
\newblock In \emph{The Eleventh International Conference on Learning Representations, {ICLR} 2023, Kigali, Rwanda, May 1-5, 2023}.

\bibitem[{Lakshminarayanan et~al.(2017)Lakshminarayanan, Pritzel, and Blundell}]{NIPS2017_9ef2ed4b}
Balaji Lakshminarayanan, Alexander Pritzel, and Charles Blundell. 2017.
\newblock \href {https://proceedings.neurips.cc/paper_files/paper/2017/file/9ef2ed4b7fd2c810847ffa5fa85bce38-Paper.pdf} {Simple and scalable predictive uncertainty estimation using deep ensembles}.
\newblock In \emph{Advances in Neural Information Processing Systems}, volume~30. Curran Associates, Inc.

\bibitem[{Lee et~al.(2018)Lee, Lee, Lee, and Shin}]{lee2018simple}
Kimin Lee, Kibok Lee, Honglak Lee, and Jinwoo Shin. 2018.
\newblock \href {https://proceedings.neurips.cc/paper/2018/hash/abdeb6f575ac5c6676b747bca8d09cc2-Abstract.html} {A simple unified framework for detecting out-of-distribution samples and adversarial attacks}.
\newblock In \emph{Advances in Neural Information Processing Systems 31: Annual Conference on Neural Information Processing Systems 2018, NeurIPS 2018, December 3-8, 2018, Montr{\'{e}}al, Canada}, volume~31, pages 7167--7177.

\bibitem[{Lin et~al.(2022)Lin, Hilton, and Evans}]{lin-etal-2022-truthfulqa}
Stephanie Lin, Jacob Hilton, and Owain Evans. 2022.
\newblock \href {https://doi.org/10.18653/v1/2022.acl-long.229} {{T}ruthful{QA}: Measuring how models mimic human falsehoods}.
\newblock In \emph{Proceedings of the 60th Annual Meeting of the Association for Computational Linguistics (Volume 1: Long Papers)}, pages 3214--3252, Dublin, Ireland. Association for Computational Linguistics.

\bibitem[{Lin et~al.(2023)Lin, Trivedi, and Sun}]{lin2023generating}
Zhen Lin, Shubhendu Trivedi, and Jimeng Sun. 2023.
\newblock \href {https://openreview.net/pdf?id=DWkJCSxKU5} {Generating with confidence: Uncertainty quantification for black-box large language models}.
\newblock \emph{Transactions on Machine Learning Research}.

\bibitem[{Liu et~al.(2020)Liu, Lin, Padhy, Tran, Bedrax~Weiss, and Lakshminarayanan}]{NEURIPS2020_543e8374}
Jeremiah Liu, Zi~Lin, Shreyas Padhy, Dustin Tran, Tania Bedrax~Weiss, and Balaji Lakshminarayanan. 2020.
\newblock \href {https://proceedings.neurips.cc/paper_files/paper/2020/file/543e83748234f7cbab21aa0ade66565f-Paper.pdf} {Simple and principled uncertainty estimation with deterministic deep learning via distance awareness}.
\newblock In \emph{Advances in Neural Information Processing Systems}, volume~33, pages 7498--7512. Curran Associates, Inc.

\bibitem[{Malinin and Gales(2021)}]{malinin21uncertainty}
Andrey Malinin and Mark J.~F. Gales. 2021.
\newblock \href {https://openreview.net/forum?id=jN5y-zb5Q7m} {Uncertainty estimation in autoregressive structured prediction}.
\newblock In \emph{9th International Conference on Learning Representations, {ICLR} 2021, Virtual Event, Austria, May 3-7, 2021}. OpenReview.net.

\bibitem[{Manakul et~al.(2023)Manakul, Liusie, and Gales}]{manakul2023selfcheckgpt}
Potsawee Manakul, Adian Liusie, and Mark Gales. 2023.
\newblock \href {https://doi.org/10.48550/arXiv.2303.08896} {{SelfCheckGPT}: Zero-resource black-box hallucination detection for generative large language models}.
\newblock In \emph{Proceedings of the 2023 Conference on Empirical Methods in Natural Language Processing}, pages 9004--9017.

\bibitem[{Min et~al.(2023)Min, Krishna, Lyu, Lewis, Yih, Koh, Iyyer, Zettlemoyer, and Hajishirzi}]{min2023factscore}
Sewon Min, Kalpesh Krishna, Xinxi Lyu, Mike Lewis, Wen-tau Yih, Pang Koh, Mohit Iyyer, Luke Zettlemoyer, and Hannaneh Hajishirzi. 2023.
\newblock \href {https://doi.org/10.18653/v1/2023.emnlp-main.741} {{FA}ct{S}core: Fine-grained atomic evaluation of factual precision in long form text generation}.
\newblock In \emph{Proceedings of the 2023 Conference on Empirical Methods in Natural Language Processing}, pages 12076--12100.

\bibitem[{Narayan et~al.(2018)Narayan, Cohen, and Lapata}]{xsum}
Shashi Narayan, Shay~B. Cohen, and Mirella Lapata. 2018.
\newblock \href {https://doi.org/10.18653/v1/d18-1206} {Don't give me the details, just the summary! topic-aware convolutional neural networks for extreme summarization}.
\newblock In \emph{Proceedings of the 2018 Conference on Empirical Methods in Natural Language Processing, Brussels, Belgium, October 31 - November 4, 2018}, pages 1797--1807. Association for Computational Linguistics.

\bibitem[{Nikitin et~al.(2024)Nikitin, Kossen, Gal, and Marttinen}]{nikitin2024kernel}
Alexander Nikitin, Jannik Kossen, Yarin Gal, and Pekka Marttinen. 2024.
\newblock \href {https://arxiv.org/abs/2405.20003} {Kernel language entropy: Fine-grained uncertainty quantification for llms from semantic similarities}.
\newblock \emph{arXiv preprint arXiv:2405.20003}.

\bibitem[{OpenAI et~al.(2024)OpenAI, Achiam, Adler, Agarwal, Ahmad, Akkaya, Aleman, Almeida, Altenschmidt, Altman, Anadkat, Avila, Babuschkin, Balaji, Balcom, Baltescu, Bao, Bavarian, Belgum, Bello, Berdine, Bernadett-Shapiro, Berner, Bogdonoff, Boiko, Boyd, Brakman, Brockman, Brooks, Brundage, Button, Cai, Campbell, Cann, Carey, Carlson, Carmichael, Chan, Chang, Chantzis, Chen, Chen, Chen, Chen, Chen, Chess, Cho, Chu, Chung, Cummings, Currier, Dai, Decareaux, Degry, Deutsch, Deville, Dhar, Dohan, Dowling, Dunning, Ecoffet, Eleti, Eloundou, Farhi, Fedus, Felix, Fishman, Forte, Fulford, Gao, Georges, Gibson, Goel, Gogineni, Goh, Gontijo-Lopes, Gordon, Grafstein, Gray, Greene, Gross, Gu, Guo, Hallacy, Han, Harris, He, Heaton, Heidecke, Hesse, Hickey, Hickey, Hoeschele, Houghton, Hsu, Hu, Hu, Huizinga, Jain, Jain, Jang, Jiang, Jiang, Jin, Jin, Jomoto, Jonn, Jun, Kaftan, Łukasz Kaiser, Kamali, Kanitscheider, Keskar, Khan, Kilpatrick, Kim, Kim, Kim, Kirchner, Kiros, Knight, Kokotajlo, Łukasz Kondraciuk,
  Kondrich, Konstantinidis, Kosic, Krueger, Kuo, Lampe, Lan, Lee, Leike, Leung, Levy, Li, Lim, Lin, Lin, Litwin, Lopez, Lowe, Lue, Makanju, Malfacini, Manning, Markov, Markovski, Martin, Mayer, Mayne, McGrew, McKinney, McLeavey, McMillan, McNeil, Medina, Mehta, Menick, and et~al.}]{openai2024gpt4technicalreport}
OpenAI, Josh Achiam, Steven Adler, Sandhini Agarwal, Lama Ahmad, Ilge Akkaya, Florencia~Leoni Aleman, Diogo Almeida, Janko Altenschmidt, Sam Altman, Shyamal Anadkat, Red Avila, Igor Babuschkin, Suchir Balaji, Valerie Balcom, Paul Baltescu, Haiming Bao, Mohammad Bavarian, Jeff Belgum, Irwan Bello, Jake Berdine, Gabriel Bernadett-Shapiro, Christopher Berner, Lenny Bogdonoff, Oleg Boiko, Madelaine Boyd, Anna-Luisa Brakman, Greg Brockman, Tim Brooks, Miles Brundage, Kevin Button, Trevor Cai, Rosie Campbell, Andrew Cann, Brittany Carey, Chelsea Carlson, Rory Carmichael, Brooke Chan, Che Chang, Fotis Chantzis, Derek Chen, Sully Chen, Ruby Chen, Jason Chen, Mark Chen, Ben Chess, Chester Cho, Casey Chu, Hyung~Won Chung, Dave Cummings, Jeremiah Currier, Yunxing Dai, Cory Decareaux, Thomas Degry, Noah Deutsch, Damien Deville, Arka Dhar, David Dohan, Steve Dowling, Sheila Dunning, Adrien Ecoffet, Atty Eleti, Tyna Eloundou, David Farhi, Liam Fedus, Niko Felix, Simón~Posada Fishman, Juston Forte, Isabella Fulford, Leo
  Gao, Elie Georges, Christian Gibson, Vik Goel, Tarun Gogineni, Gabriel Goh, Rapha Gontijo-Lopes, Jonathan Gordon, Morgan Grafstein, Scott Gray, Ryan Greene, Joshua Gross, Shixiang~Shane Gu, Yufei Guo, Chris Hallacy, Jesse Han, Jeff Harris, Yuchen He, Mike Heaton, Johannes Heidecke, Chris Hesse, Alan Hickey, Wade Hickey, Peter Hoeschele, Brandon Houghton, Kenny Hsu, Shengli Hu, Xin Hu, Joost Huizinga, Shantanu Jain, Shawn Jain, Joanne Jang, Angela Jiang, Roger Jiang, Haozhun Jin, Denny Jin, Shino Jomoto, Billie Jonn, Heewoo Jun, Tomer Kaftan, Łukasz Kaiser, Ali Kamali, Ingmar Kanitscheider, Nitish~Shirish Keskar, Tabarak Khan, Logan Kilpatrick, Jong~Wook Kim, Christina Kim, Yongjik Kim, Jan~Hendrik Kirchner, Jamie Kiros, Matt Knight, Daniel Kokotajlo, Łukasz Kondraciuk, Andrew Kondrich, Aris Konstantinidis, Kyle Kosic, Gretchen Krueger, Vishal Kuo, Michael Lampe, Ikai Lan, Teddy Lee, Jan Leike, Jade Leung, Daniel Levy, Chak~Ming Li, Rachel Lim, Molly Lin, Stephanie Lin, Mateusz Litwin, Theresa Lopez, Ryan
  Lowe, Patricia Lue, Anna Makanju, Kim Malfacini, Sam Manning, Todor Markov, Yaniv Markovski, Bianca Martin, Katie Mayer, Andrew Mayne, Bob McGrew, Scott~Mayer McKinney, Christine McLeavey, Paul McMillan, Jake McNeil, David Medina, Aalok Mehta, Jacob Menick, and Luke~Metz et~al. 2024.
\newblock \href {https://arxiv.org/abs/2303.08774} {Gpt-4 technical report}.
\newblock \emph{Preprint}, arXiv:2303.08774.

\bibitem[{Podolskiy et~al.(2021)Podolskiy, Lipin, Bout, Artemova, and Piontkovskaya}]{podolskiy2021revisiting}
Alexander Podolskiy, Dmitry Lipin, Andrey Bout, Ekaterina Artemova, and Irina Piontkovskaya. 2021.
\newblock \href {https://cdn.aaai.org/ojs/17612/17612-13-21106-1-2-20210518.pdf} {Revisiting mahalanobis distance for transformer-based out-of-domain detection}.
\newblock In \emph{Proceedings of the AAAI Conference on Artificial Intelligence}, pages 13675--13682.

\bibitem[{Raffel et~al.(2020)Raffel, Shazeer, Roberts, Lee, Narang, Matena, Zhou, Li, and Liu}]{c4dataset}
Colin Raffel, Noam Shazeer, Adam Roberts, Katherine Lee, Sharan Narang, Michael Matena, Yanqi Zhou, Wei Li, and Peter~J. Liu. 2020.
\newblock \href {https://jmlr.org/papers/v21/20-074.html} {Exploring the limits of transfer learning with a unified text-to-text transformer}.
\newblock \emph{J. Mach. Learn. Res.}, 21:140:1--140:67.

\bibitem[{Reddy et~al.(2019)Reddy, Chen, and Manning}]{coqa}
Siva Reddy, Danqi Chen, and Christopher~D. Manning. 2019.
\newblock \href {https://doi.org/10.1162/tacl_a_00266} {{C}o{QA}: A conversational question answering challenge}.
\newblock \emph{Transactions of the Association for Computational Linguistics}, 7:249--266.

\bibitem[{Ren et~al.(2023)Ren, Luo, Zhao, Krishna, Saleh, Lakshminarayanan, and Liu}]{ren2023outofdistribution}
Jie Ren, Jiaming Luo, Yao Zhao, Kundan Krishna, Mohammad Saleh, Balaji Lakshminarayanan, and Peter~J Liu. 2023.
\newblock \href {https://openreview.net/forum?id=kJUS5nD0vPB} {Out-of-distribution detection and selective generation for conditional language models}.
\newblock In \emph{The Eleventh International Conference on Learning Representations}.

\bibitem[{Rivi{\`{e}}re et~al.(2024)Rivi{\`{e}}re, Pathak, Sessa, Hardin, Bhupatiraju, Hussenot, Mesnard, Shahriari, Ram{\'{e}}, Ferret, Liu, Tafti, Friesen, Casbon, Ramos, Kumar, Lan, Jerome, Tsitsulin, Vieillard, Stanczyk, Girgin, Momchev, Hoffman, Thakoor, Grill, Neyshabur, Bachem, Walton, Severyn, Parrish, Ahmad, Hutchison, Abdagic, Carl, Shen, Brock, Coenen, Laforge, Paterson, Bastian, Piot, Wu, Royal, Chen, Kumar, Perry, Welty, Choquette{-}Choo, Sinopalnikov, Weinberger, Vijaykumar, Rogozinska, Herbison, Bandy, Wang, Noland, Moreira, Senter, Eltyshev, Visin, Rasskin, Wei, Cameron, Martins, Hashemi, Klimczak{-}Plucinska, Batra, Dhand, Nardini, Mein, Zhou, Svensson, Stanway, Chan, Zhou, Carrasqueira, Iljazi, Becker, Fernandez, van Amersfoort, Gordon, Lipschultz, Newlan, Ji, Mohamed, Badola, Black, Millican, McDonell, Nguyen, Sodhia, Greene, Sj{\"{o}}sund, Usui, Sifre, Heuermann, Lago, and McNealus}]{gemma2}
Morgane Rivi{\`{e}}re, Shreya Pathak, Pier~Giuseppe Sessa, Cassidy Hardin, Surya Bhupatiraju, L{\'{e}}onard Hussenot, Thomas Mesnard, Bobak Shahriari, Alexandre Ram{\'{e}}, Johan Ferret, Peter Liu, Pouya Tafti, Abe Friesen, Michelle Casbon, Sabela Ramos, Ravin Kumar, Charline~Le Lan, Sammy Jerome, Anton Tsitsulin, Nino Vieillard, Piotr Stanczyk, Sertan Girgin, Nikola Momchev, Matt Hoffman, Shantanu Thakoor, Jean{-}Bastien Grill, Behnam Neyshabur, Olivier Bachem, Alanna Walton, Aliaksei Severyn, Alicia Parrish, Aliya Ahmad, Allen Hutchison, Alvin Abdagic, Amanda Carl, Amy Shen, Andy Brock, Andy Coenen, Anthony Laforge, Antonia Paterson, Ben Bastian, Bilal Piot, Bo~Wu, Brandon Royal, Charlie Chen, Chintu Kumar, Chris Perry, Chris Welty, Christopher~A. Choquette{-}Choo, Danila Sinopalnikov, David Weinberger, Dimple Vijaykumar, Dominika Rogozinska, Dustin Herbison, Elisa Bandy, Emma Wang, Eric Noland, Erica Moreira, Evan Senter, Evgenii Eltyshev, Francesco Visin, Gabriel Rasskin, Gary Wei, Glenn Cameron, Gus
  Martins, Hadi Hashemi, Hanna Klimczak{-}Plucinska, Harleen Batra, Harsh Dhand, Ivan Nardini, Jacinda Mein, Jack Zhou, James Svensson, Jeff Stanway, Jetha Chan, Jin~Peng Zhou, Joana Carrasqueira, Joana Iljazi, Jocelyn Becker, Joe Fernandez, Joost van Amersfoort, Josh Gordon, Josh Lipschultz, Josh Newlan, Ju{-}yeong Ji, Kareem Mohamed, Kartikeya Badola, Kat Black, Katie Millican, Keelin McDonell, Kelvin Nguyen, Kiranbir Sodhia, Kish Greene, Lars~Lowe Sj{\"{o}}sund, Lauren Usui, Laurent Sifre, Lena Heuermann, Leticia Lago, and Lilly McNealus. 2024.
\newblock \href {https://doi.org/10.48550/arXiv.2408.00118} {Gemma 2: Improving open language models at a practical size}.
\newblock \emph{arXiv preprint arXiv:2408.00118}.

\bibitem[{Rousseeuw(1984)}]{Rousseeuw84leastmedian}
Peter~J Rousseeuw. 1984.
\newblock \href {https://doi.org/10.1080/01621459.1984.10477105} {Least median of squares regression}.
\newblock \emph{Journal of the American statistical association}, 79(388):871--880.

\bibitem[{See et~al.(2017)See, Liu, and Manning}]{see-etal-2017-get}
Abigail See, Peter~J. Liu, and Christopher~D. Manning. 2017.
\newblock \href {https://doi.org/10.18653/v1/P17-1099} {Get to the point: Summarization with pointer-generator networks}.
\newblock In \emph{Proceedings of the 55th Annual Meeting of the Association for Computational Linguistics (Volume 1: Long Papers)}, pages 1073--1083, Vancouver, Canada. Association for Computational Linguistics.

\bibitem[{Shelmanov et~al.(2021)Shelmanov, Tsymbalov, Puzyrev, Fedyanin, Panchenko, and Panov}]{shelmanov-etal-2021-certain}
Artem Shelmanov, Evgenii Tsymbalov, Dmitri Puzyrev, Kirill Fedyanin, Alexander Panchenko, and Maxim Panov. 2021.
\newblock \href {https://www.aclweb.org/anthology/2021.eacl-main.157} {How certain is your transformer?}
\newblock In \emph{Proceedings of the 16th Conference of the European Chapter of the Association for Computational Linguistics: Main Volume}, pages 1833--1840, Online. Association for Computational Linguistics.

\bibitem[{Shrestha(2020)}]{multicollinearity}
Noora Shrestha. 2020.
\newblock \href {https://doi.org/10.12691/ajams-8-2-1} {Detecting multicollinearity in regression analysis}.
\newblock \emph{American Journal of Applied Mathematics and Statistics}, 8:39--42.

\bibitem[{Tsymbalov et~al.(2018)Tsymbalov, Panov, and Shapeev}]{tsymbalov2018dropout}
Evgenii Tsymbalov, Maxim Panov, and Alexander Shapeev. 2018.
\newblock \href {https://link.springer.com/chapter/10.1007/978-3-030-11027-7_24} {Dropout-based active learning for regression}.
\newblock In \emph{Analysis of Images, Social Networks and Texts: 7th International Conference, AIST 2018, Moscow, Russia, July 5--7, 2018}, pages 247--258.

\bibitem[{van Amersfoort et~al.(2020)van Amersfoort, Smith, Teh, and Gal}]{ddu_amersfoort}
Joost van Amersfoort, Lewis Smith, Yee~Whye Teh, and Yarin Gal. 2020.
\newblock \href {http://proceedings.mlr.press/v119/van-amersfoort20a.html} {Uncertainty estimation using a single deep deterministic neural network}.
\newblock In \emph{Proceedings of the 37th International Conference on Machine Learning, {ICML} 2020, 13-18 July 2020, Virtual Event}, volume 119 of \emph{Proceedings of Machine Learning Research}, pages 9690--9700. {PMLR}.

\bibitem[{Vashurin et~al.(2024)Vashurin, Fadeeva, Vazhentsev, Rvanova, Tsvigun, Vasilev, Xing, Sadallah, Grishchenkov, Petrakov, Panchenko, Baldwin, Nakov, Panov, and Shelmanov}]{vashurin2024benchmakring}
Roman Vashurin, Ekaterina Fadeeva, Artem Vazhentsev, Lyudmila Rvanova, Akim Tsvigun, Daniil Vasilev, Rui Xing, Abdelrahman~Boda Sadallah, Kirill Grishchenkov, Sergey Petrakov, Alexander Panchenko, Timothy Baldwin, Preslav Nakov, Maxim Panov, and Artem Shelmanov. 2024.
\newblock \href {https://arxiv.org/abs/2406.15627} {Benchmarking uncertainty quantification methods for large language models with lm-polygraph}.
\newblock \emph{arXiv preprint arXiv:2406.15627}.

\bibitem[{Vashurin et~al.(2025)Vashurin, Goloburda, Nakov, Shelmanov, and Panov}]{vashurin2025cocoageneralizedapproachuncertainty}
Roman Vashurin, Maiya Goloburda, Preslav Nakov, Artem Shelmanov, and Maxim Panov. 2025.
\newblock \href {https://arxiv.org/abs/2502.04964} {Cocoa: A generalized approach to uncertainty quantification by integrating confidence and consistency of llm outputs}.
\newblock \emph{arXiv preprint arXiv:2502.04964}.

\bibitem[{Vazhentsev et~al.(2024)Vazhentsev, Fadeeva, Xing, Panchenko, Nakov, Baldwin, Panov, and Shelmanov}]{vazhentsev2024unconditional}
Artem Vazhentsev, Ekaterina Fadeeva, Rui Xing, Alexander Panchenko, Preslav Nakov, Timothy Baldwin, Maxim Panov, and Artem Shelmanov. 2024.
\newblock \href {https://arxiv.org/abs/2408.10692} {Unconditional truthfulness: Learning conditional dependency for uncertainty quantification of large language models}.
\newblock \emph{arXiv preprint arXiv:2408.10692}.

\bibitem[{Vazhentsev et~al.(2022)Vazhentsev, Kuzmin, Shelmanov, Tsvigun, Tsymbalov, Fedyanin, Panov, Panchenko, Gusev, Burtsev, Avetisian, and Zhukov}]{vazhentsev-etal-2022-uncertainty}
Artem Vazhentsev, Gleb Kuzmin, Artem Shelmanov, Akim Tsvigun, Evgenii Tsymbalov, Kirill Fedyanin, Maxim Panov, Alexander Panchenko, Gleb Gusev, Mikhail Burtsev, Manvel Avetisian, and Leonid Zhukov. 2022.
\newblock \href {https://doi.org/10.18653/v1/2022.acl-long.566} {Uncertainty estimation of transformer predictions for misclassification detection}.
\newblock In \emph{Proceedings of the 60th Annual Meeting of the Association for Computational Linguistics (Volume 1: Long Papers)}, pages 8237--8252, Dublin, Ireland. Association for Computational Linguistics.

\bibitem[{Vazhentsev et~al.(2023{\natexlab{a}})Vazhentsev, Kuzmin, Tsvigun, Panchenko, Panov, Burtsev, and Shelmanov}]{vazhentsev-etal-2023-hybrid}
Artem Vazhentsev, Gleb Kuzmin, Akim Tsvigun, Alexander Panchenko, Maxim Panov, Mikhail Burtsev, and Artem Shelmanov. 2023{\natexlab{a}}.
\newblock \href {https://aclanthology.org/2023.acl-long.652} {Hybrid uncertainty quantification for selective text classification in ambiguous tasks}.
\newblock In \emph{Proceedings of the 61st Annual Meeting of the Association for Computational Linguistics (Volume 1: Long Papers)}, pages 11659--11681, Toronto, Canada. Association for Computational Linguistics.

\bibitem[{Vazhentsev et~al.(2023{\natexlab{b}})Vazhentsev, Tsvigun, Vashurin, Petrakov, Vasilev, Panov, Panchenko, and Shelmanov}]{vazhentsev-etal-2023-efficient}
Artem Vazhentsev, Akim Tsvigun, Roman Vashurin, Sergey Petrakov, Daniil Vasilev, Maxim Panov, Alexander Panchenko, and Artem Shelmanov. 2023{\natexlab{b}}.
\newblock \href {https://aclanthology.org/2023.findings-acl.93} {Efficient out-of-domain detection for sequence to sequence models}.
\newblock In \emph{Findings of the Association for Computational Linguistics: ACL 2023}, pages 1430--1454, Toronto, Canada. Association for Computational Linguistics.

\bibitem[{Wang et~al.(2022)Wang, Beck, Baldwin, and Verspoor}]{wang-etal-2022-uncertainty}
Yuxia Wang, Daniel Beck, Timothy Baldwin, and Karin Verspoor. 2022.
\newblock \href {https://doi.org/10.1162/tacl_a_00483} {Uncertainty estimation and reduction of pre-trained models for text regression}.
\newblock \emph{Transactions of the Association for Computational Linguistics}, 10:680--696.

\bibitem[{Welbl et~al.(2017)Welbl, Liu, and Gardner}]{welbl-etal-2017-crowdsourcing}
Johannes Welbl, Nelson~F. Liu, and Matt Gardner. 2017.
\newblock \href {https://doi.org/10.18653/v1/W17-4413} {Crowdsourcing multiple choice science questions}.
\newblock In \emph{Proceedings of the 3rd Workshop on Noisy User-generated Text}, pages 94--106, Copenhagen, Denmark. Association for Computational Linguistics.

\bibitem[{Xiao and Wang(2021)}]{xiao-wang-2021-hallucination}
Yijun Xiao and William~Yang Wang. 2021.
\newblock \href {https://doi.org/10.18653/v1/2021.eacl-main.236} {On hallucination and predictive uncertainty in conditional language generation}.
\newblock In \emph{Proceedings of the 16th Conference of the European Chapter of the Association for Computational Linguistics: Main Volume}, pages 2734--2744, Online. Association for Computational Linguistics.

\bibitem[{Xin et~al.(2021)Xin, Tang, Yu, and Lin}]{xin-etal-2021-art}
Ji~Xin, Raphael Tang, Yaoliang Yu, and Jimmy Lin. 2021.
\newblock \href {https://doi.org/10.18653/v1/2021.acl-long.84} {The art of abstention: Selective prediction and error regularization for natural language processing}.
\newblock In \emph{Proceedings of the 59th Annual Meeting of the Association for Computational Linguistics and the 11th International Joint Conference on Natural Language Processing (Volume 1: Long Papers)}, pages 1040--1051, Online. Association for Computational Linguistics.

\bibitem[{Yoo et~al.(2022)Yoo, Kim, Jang, and Kwak}]{yoo-etal-2022-detection}
KiYoon Yoo, Jangho Kim, Jiho Jang, and Nojun Kwak. 2022.
\newblock \href {https://doi.org/10.18653/v1/2022.findings-acl.289} {Detection of adversarial examples in text classification: Benchmark and baseline via robust density estimation}.
\newblock In \emph{Findings of the Association for Computational Linguistics: ACL 2022}, pages 3656--3672, Dublin, Ireland. Association for Computational Linguistics.

\bibitem[{Zha et~al.(2023)Zha, Yang, Li, and Hu}]{zha-etal-2023-alignscore}
Yuheng Zha, Yichi Yang, Ruichen Li, and Zhiting Hu. 2023.
\newblock \href {https://doi.org/10.18653/v1/2023.acl-long.634} {{A}lign{S}core: Evaluating factual consistency with a unified alignment function}.
\newblock In \emph{Proceedings of the 61st Annual Meeting of the Association for Computational Linguistics (Volume 1: Long Papers)}, pages 11328--11348, Toronto, Canada. Association for Computational Linguistics.

\bibitem[{Zhang et~al.(2019)Zhang, Chen, Lu, and Ramakrishnan}]{zhang-etal-2019-mitigating}
Xuchao Zhang, Fanglan Chen, Chang-Tien Lu, and Naren Ramakrishnan. 2019.
\newblock \href {https://doi.org/10.18653/v1/N19-1316} {Mitigating uncertainty in document classification}.
\newblock In \emph{Proceedings of the 2019 Conference of the North {A}merican Chapter of the Association for Computational Linguistics: Human Language Technologies, Volume 1 (Long and Short Papers)}, pages 3126--3136, Minneapolis, Minnesota. Association for Computational Linguistics.

\end{thebibliography}
